\documentclass{article}

\usepackage[final,nonatbib]{neurips_data_2023}

\usepackage[utf8]{inputenc} % allow utf-8 input
\usepackage[T1]{fontenc}    % use 8-bit T1 fonts
\usepackage{hyperref}       % hyperlinks
\usepackage{url}            % simple URL typesetting
\usepackage{booktabs}       % professional-quality tables
\usepackage{amsfonts}       % blackboard math symbols
\usepackage{nicefrac}       % compact symbols for 1/2, etc.
\usepackage{microtype}      % microtypography
\usepackage[table,xcdraw]{xcolor}         % colors
\usepackage{graphicx}
\usepackage{enumitem}
\usepackage{amsmath}
\usepackage{multirow}
\usepackage{listings}

\hypersetup{
  colorlinks=true,
  filecolor=magenta,
}

\title{OpenSTL: A Comprehensive Benchmark of Spatio-Temporal Predictive Learning}

\author{%
Cheng Tan$^{1,2*}$\quad Siyuan Li$^{1,2}$\thanks{Equal contribution.}\quad Zhangyang Gao$^{1,2}$\quad Wenfei Guan$^{3}$\quad Zedong Wang$^{2}$\\ 
\textbf{Zicheng Liu$^{1,2}$\quad Lirong Wu$^{1,2}$\quad \ Stan Z. Li}$^{2}$\thanks{Corresponding author.} \\
$^{1}$Zhejiang University; \quad\quad $^{3}$Xidian University; \\
$^{2}$AI Lab, Research Center for Industries of the Future, Westlake University\\
\vspace{0.25em}
\texttt{\{tancheng; lisiyuan; gaozhangyang; guanwenfei; wangzedong; liuzicheng;}\\
\texttt{wulirong; stan.zq.li\}@westlake.edu.cn};
}

\begin{document}

\maketitle

\begin{abstract}
Spatio-temporal predictive learning is a learning paradigm that enables models to learn spatial and temporal patterns by predicting future frames from given past frames in an unsupervised manner. Despite remarkable progress in recent years, a lack of systematic understanding persists due to the diverse settings, complex implementation, and difficult reproducibility. Without standardization, comparisons can be unfair and insights inconclusive. To address this dilemma, we propose OpenSTL, a comprehensive benchmark for spatio-temporal predictive learning that categorizes prevalent approaches into recurrent-based and recurrent-free models. OpenSTL provides a modular and extensible framework implementing various state-of-the-art methods. We conduct standard evaluations on datasets across various domains, including synthetic moving object trajectory, human motion, driving scenes, traffic flow, and weather forecasting. Based on our observations, we provide a detailed analysis of how model architecture and dataset properties affect spatio-temporal predictive learning performance. Surprisingly, we find that recurrent-free models achieve a good balance between efficiency and performance than recurrent models. Thus, we further extend the common MetaFormers to boost recurrent-free spatial-temporal predictive learning. We open-source the code and models at \href{https://github.com/chengtan9907/OpenSTL}{https://github.com/chengtan9907/OpenSTL}.
\end{abstract}
% answer a long-standing question in this field: 

\section{Introduction}

Recent years have witnessed rapid and remarkable progress in spatio-temporal predictive learning~\cite{convlstm,prednet,simvp,tan2022temporal}. This burgeoning field aims to learn latent spatial and temporal patterns through the challenging task of forecasting future frames based solely on given past frames in an unsupervised manner~\cite{tan2022simvp,wu2022knowledge,wu2021self,wu2022graphmixup}. By ingesting raw sequential data, these self-supervised models~\cite{icml2020simclr,cvpr2022mae, icml2023a2mim} can uncover intricate spatial and temporal interdependencies without the need for tedious manual annotation, enabling them to extrapolate coherently into the future in a realistic fashion~\cite{prednet,phydnet}. Spatio-temporal predictive learning benefits a wide range of applications with its ability to anticipate the future from the past in a data-driven way, including modeling the devastating impacts of climate change~\cite{convlstm,reichstein2019deep}, predicting human movement~\cite{zhang2017learning,wang2018rgb}, forecasting traffic flow in transportation systems~\cite{fang2019gstnet,mim}, and learning expressive representations from video~\cite{qian2021spatiotemporal,jenni2020video}. By learning to predict the future without supervision from massive datasets, these techniques have the potential to transform domains where anticipation and planning are crucial but limited labeled data exists~\cite{finn2016unsupervised,castrejon2019improved,villegas2018hierarchical,oprea2020review}.

\begin{figure}[ht]
  \centering
  \includegraphics[width=0.98\textwidth]{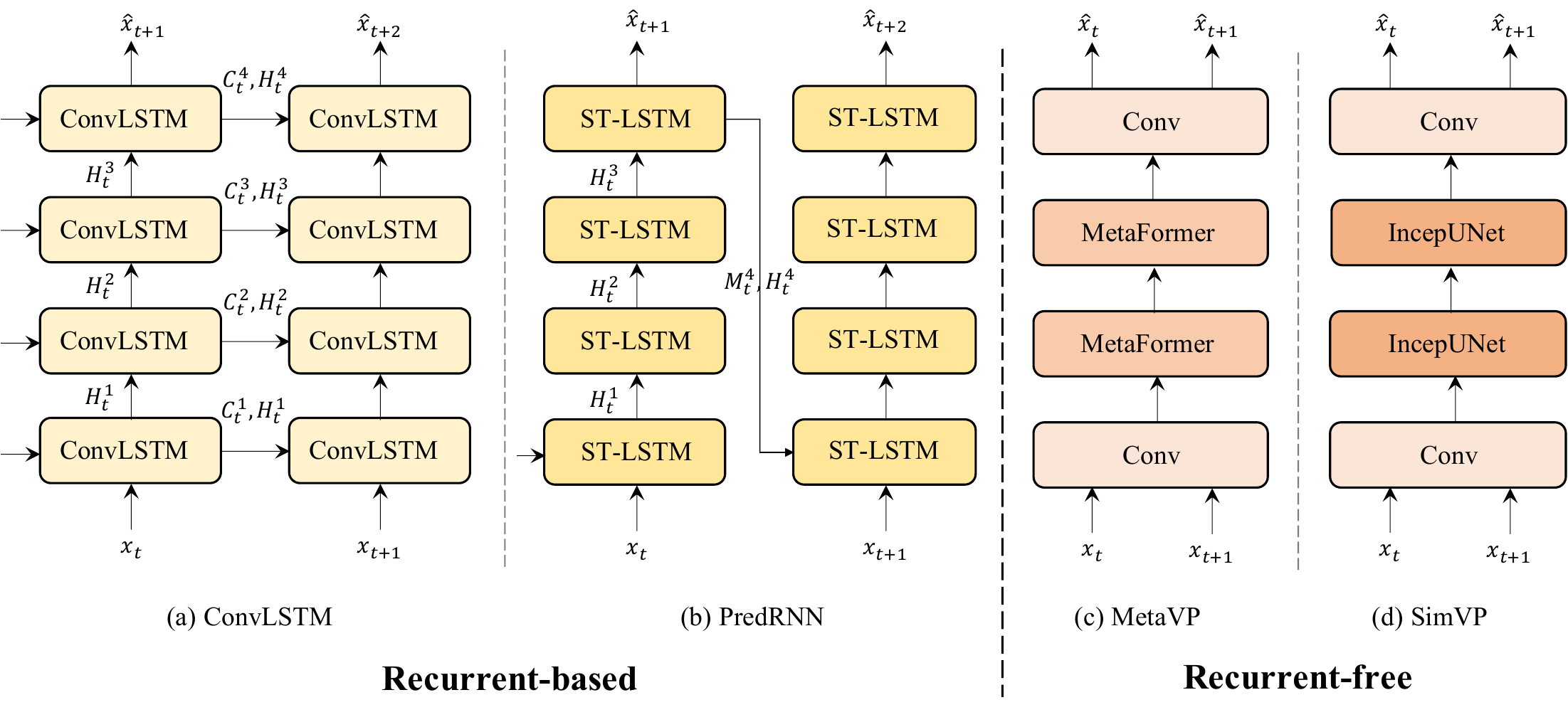}
  \vspace{-1.0em}
  \caption{Two typical sptaio-temporal predictive learning models. As illustrated by the left two instances (a)(b), the first type requires several recurrent modules to predict the next frame according to the previous frames in an auto-regressive manner, dubbed recurrent-based models. As for the right two instances (c)(d), the second type predicts all future frames based on all given frames at once, which inferences in parallel and is called the recurrent-free model.}
  \label{fig:intro_comparison}
\end{figure}

Despite the significance of spatio-temporal predictive learning and the development of various approaches, there remains a conspicuous lack of a comprehensive benchmark for this field covering various synthetic and practical application scenarios. We believe that a comprehensive benchmark is essential for advancing the field and facilitating meaningful comparisons between different methods. In particular, there exists a perennial question that has not yet been conclusively answered: \textit{is it necessary to employ recurrent neural network architectures to capture temporal dependencies?} In other words, \textit{can recurrent-free models achieve performance comparable to recurrent-based models without explicit temporal modeling}? 

Since the seminal work ConvLSTM~\cite{convlstm} was proposed, which ingeniously integrates convolutional networks and long-short term memory (LSTM) networks~\cite{hochreiter1997long} to separately capture spatial and temporal correlations, researchers have vacillated between utilizing or eschewing recurrent architectures. As shown in Figure~\ref{fig:intro_comparison}, (a) ConvLSTM is a prototypical recurrent-based model that infuses a recurrent structure into convolutional networks. (b) PredRNN~\cite{predrnn} represents a series of recurrent models that revise the flow of information to enhance performance. (c) MetaVP is the recurrent-free model that abstracted from SimVP by substituting its IncepU~\cite{simvp} modules with MetaFormers~\cite{yu2022metaformer}. (d) SimVP~\cite{simvp,tan2022simvp} is a typical recurrent-free model that achieves performance comparable to previous state-of-the-art models without explicitly modeling temporal dependencies. 

In this study, we illuminate the long-standing question of whether explicit temporal modeling with recurrent neural networks is requisite for spatio-temporal predictive learning. To achieve this, we present a comprehensive benchmark, \textbf{Open} \textbf{S}patio-\textbf{T}emporal predictive \textbf{L}earning, dubbed OpenSTL. We revisit the approaches that represent the foremost strands within a modular and extensive framework to ensure fair comparisons. We summarize our main contributions as follows:

\begin{itemize}
  \item We build OpenSTL, a comprehensive benchmark for spatio-temporal predictive learning that includes 14 representative algorithms and 24 models. OpenSTL covers a wide range of methods and classifies them into two categories: recurrent-based and recurrent-free methods. 
  \item We conduct extensive experiments on a diversity of tasks ranging from synthetic moving object trajectories to real-world human motion, driving scenes, traffic flow, and weather forecasting. The datasets span synthetic to real-world data and micro-to-macro scales.
  \item While recurrent-based models have been well developed, we rethink the potential of recurrent-free models based on insights from OpenSTL. We propose generalizing MetaFormer-like architectures~\cite{yu2022metaformer} to boost recurrent-free spatio-temporal predictive learning. Recurrent-free models can thus reformulate the problem as a downstream task of designing vision backbones for general applications.
\end{itemize}

\section{Background and Related work}

\subsection{Problem definition}
% \vspace{-2mm}

We propose the formal definition for the spatio-temporal predictive learning problem as follows. Given a sequence of video frames $\mathcal{X}^{t, T} = \{\boldsymbol{x}^i\}_{t-T+1}^t$ up to time $t$ spanning the past $T$ frames, the objective is to predict the subsequent $T'$ frames $\mathcal{Y}^{t+1, T'} = \{\boldsymbol{x}^{i}\}_{t+1}^{t+1+T'}$ from time $t+1$ onwards, where each frame $\boldsymbol{x}_i \in \mathbb{R}^{C \times H \times W}$ typically comprises $C$ channels, with height $H$ and width $W$ pixels. In practice, we represent the input sequence of observed frames and output sequence of predicted frames respectively as tensors $\mathcal{X}^{t, T} \in \mathbb{R}^{T \times C \times H \times W}$ and $\mathcal{Y}^{t+1, T'} \in \mathbb{R}^{T' \times C \times H \times W}$. 

The model with learnable parameters $\Theta$ learns a mapping $\mathcal{F}_\Theta: \mathcal{X}^{t, T} \mapsto \mathcal{Y}^{t+1, T'}$ by leveraging both spatial and temporal dependencies. In our case, the mapping $\mathcal{F}_\Theta$ corresponds to a neural network trained to minimize the discrepancy between the predicted future frames and the ground-truth future frames. The optimal parameters $\Theta^*$ are given by:
\begin{equation}
  \Theta^* = \arg\min_{\Theta} \mathcal{L}(\mathcal{F}_\Theta(\mathcal{X}^{t, T}), \mathcal{Y}^{t+1, T'}),
\end{equation}
where $\mathcal{L}$ denotes a loss function that quantifies such discrepancy.

In this study, we categorize prevalent spatio-temporal predictive learning methods into two classes: recurrent-based and recurrent-free models. For \textit{recurrent-based models}, the mapping $\mathcal{F}_\Theta$ comprises several recurrent interactions:
\begin{equation}
  \mathcal{F}_\Theta: f_\theta(\boldsymbol{x}^{t-T+1}, \boldsymbol{h}^{t-T+1}) \circ ... \circ f_\theta(\boldsymbol{x}^{t}, \boldsymbol{h}^{t}) \circ ... \circ f_\theta(\boldsymbol{x}^{t+T'-1}, \boldsymbol{h}^{t+T'-1}),
\end{equation}
where $\boldsymbol{h}^i$ represents the memory state encompassing historical information and $f_\theta$ denotes the mapping between each pair of adjacent frames. The parameters $\theta$ are shared across each state. Therefore, the prediction process can be expressed as follows:
\begin{equation}
  \boldsymbol{x}^{t+1} = f_\theta(\boldsymbol{x}^i, \boldsymbol{h}^{i}), \forall i \in \{t+1, \cdots, t+T'\},
\end{equation}
For \textit{recurrent-free} models, the prediction process directly feeds the whole sequence of observed frames into the model and outputs the complete predicted frames at once.

\subsection{Recurrent-based models}
\vspace{-1mm}

Since the pioneering work ConvLSTM~\cite{convlstm} was proposed, recurrent-based models~\cite{prednet,oliu2018folded,ddpae,phydnet,crevnet,oprea2020review} have been extensively studied. PredRNN~\cite{predrnn} adopts vanilla ConvLSTM modules to build a Spatio-temporal LSTM (ST-LSTM) unit that models spatial and temporal variations simultaneously. PredRNN++~\cite{predrnn++} proposes a gradient highway unit to mitigate the gradient vanishing and a Casual-LSTM module to cascadely connect spatial and temporal memories. PredRNNv2~\cite{predrnnv2} further proposes a curriculum learning strategy and a memory decoupling loss to boost performance. MIM~\cite{mim} introduces high-order non-stationarity learning in designing LSTM modules. PhyDNet~\cite{phydnet} explicitly disentangles PDE dynamics from unknown complementary information with a recurrent physical unit. E3DLSTM~\cite{e3dlstm} integrates 3D convolutions into recurrent networks. MAU~\cite{chang2021mau} proposes a motion-aware unit that captures motion information. Although various recurrent-based models have been developed, the reasons behind their strong performance remain not fully understood.

\subsection{Recurrent-free models}
\vspace{-1mm}

Compared to recurrent-based models, recurrent-free models have received less attention. Previous studies tend to use 3D convolutional networks to model temporal dependencies~\cite{liu2017video,aigner2018futuregan}. PredCNN~\cite{predcnn} and TrajectoryCNN~\cite{liu2020trajectorycnn} use 2D convolutional networks for efficiency. However, early recurrent-free models were doubted due to their poor performance. Recently, SimVP~\cite{simvp,tan2022simvp,tan2022temporal} provided a simple but effective recurrent-free baseline with competitive performance. PastNet~\cite{Wu2023PastNet} and IAM4VP~\cite{Seo2023Implicit} are two recent recurrent-free models that perform strong performance. In this study, we implemented representative recurrent-based and recurrent-free models under a unified framework to systematically investigate their intrinsic properties. Moreover, we further explored the potential of recurrent-free models by reformulating the spatio-temporal predictive learning problem and extending MetaFormers~\cite{yu2022metaformer} to bridge the gap between the visual backbone and spatio-temporal learning.

\section{OpenSTL}

\subsection{Supported Methods}

\subsubsection{Overview}

OpenSTL has implemented 14 representative spatio-temporal predictive learning methods under a unified framework, including 11 recurrent-based methods and 3 recurrent-free methods. We summarize these methods in Table~\ref{tab:overview}, where we also provide the corresponding conference/journal and the types of their spatial-temporal modeling components. The spatial modeling of these methods is fundamentally consistent. Most methods apply two-dimensional convolutional networks (Conv2D) to model spatial dependencies, while E3D-LSTM and CrevNet harness three-dimensional convolutional networks (Conv3D) instead.

The primary distinction between these methods lies in how they model temporal dependencies using their proposed modules. The ST-LSTM module, proposed in PredRNN~\cite{predrnn}, is the most widely used module. CrevNet has a similar modeling approach as PredRNN, but it incorporates an information-preserving mechanism into the model. Analogously, Casual-LSTM~\cite{predrnn++}, MIM Block~\cite{mim}, E3D-LSTM~\cite{e3dlstm}, PhyCell~\cite{phydnet}, and MAU~\cite{chang2021mau} represent variants of ConvLSTM proposed with miscellaneous motivations. MVFB is built as a multi-scale voxel flow block that diverges from ConvLSTM. However, DMVFN~\cite{hu2023dmvfn} predicts future frames frame-by-frame which still qualifies as a recurrent-based model. IncepU~\cite{simvp} constitutes an Unet-like module that also exploits the multi-scale feature from the InceptionNet-like architecture. gSTA~\cite{tan2022simvp} and TAU~\cite{tan2022temporal} extend the IncepU module to simpler and more efficient architectures without InceptionNet or Unet-like architectures. In this work, we further extend the temporal modeling of recurrent-free models by introducing MetaFormers~\cite{yu2022metaformer} to boost recurrent-free spatio-temporal predictive learning.

\begin{table}[ht]
\centering
\caption{Categorizations of the supported spatial-temporal predictive learning methods in OpenSTL.}
\setlength{\tabcolsep}{1mm}{
\begin{tabular}{ccccc}
\toprule
Category    & Method    & Conference/Journal & Spatial modeling & Temporal modeling \\
\midrule
\multirow{9}{*}{Recurrent-based} 
& ConvLSTM~\cite{convlstm}   & NeurIPS 2015       & Conv2D           & Conv-LSTM \\
& PredNet~\cite{prednet} & ICLR 2017          & Conv2D           & ST-LSTM           \\
& PredRNN~\cite{predrnn}     & NeurIPS 2017       & Conv2D           & ST-LSTM           \\
& PredRNN++~\cite{predrnn++} & ICML 2018          & Conv2D           & Casual-LSTM       \\
& MIM~\cite{mim}             & CVPR 2019          & Conv2D           & MIM Block         \\
& E3D-LSTM~\cite{e3dlstm}    & ICLR 2019          & Conv3D           & E3D-LSTM          \\
& CrevNet~\cite{crevnet}     & ICLR 2020          & Conv3D           & ST-LSTM           \\
& PhyDNet~\cite{phydnet}     & CVPR 2020          & Conv2D           & ConvLSTM+PhyCell  \\
& MAU~\cite{chang2021mau}    & NeurIPS 2021       & Conv2D           & MAU               \\
& PredRNNv2~\cite{predrnnv2} & TPAMI 2022         & Conv2D           & ST-LSTM           \\
& DMVFN~\cite{hu2023dmvfn} & CVPR 2023 & Conv2D & MVFB \\ 
\hline
\multirow{3}{*}{Recurrent-free}  
& SimVP~\cite{simvp} & CVPR 2022          & Conv2D & IncepU \\
& TAU~\cite{tan2022temporal}    & CVPR 2023          & Conv2D & TAU    \\
& SimVPv2~\cite{tan2022simvp}   & arXiv              & Conv2D & gSTA   \\     
\bottomrule    
\end{tabular}}
\label{tab:overview}
\end{table}

\subsubsection{Rethink the recurrent-free models}

\begin{figure}[ht]
  \centering
  \vspace{-0.5em}
  \includegraphics[width=0.93\textwidth]{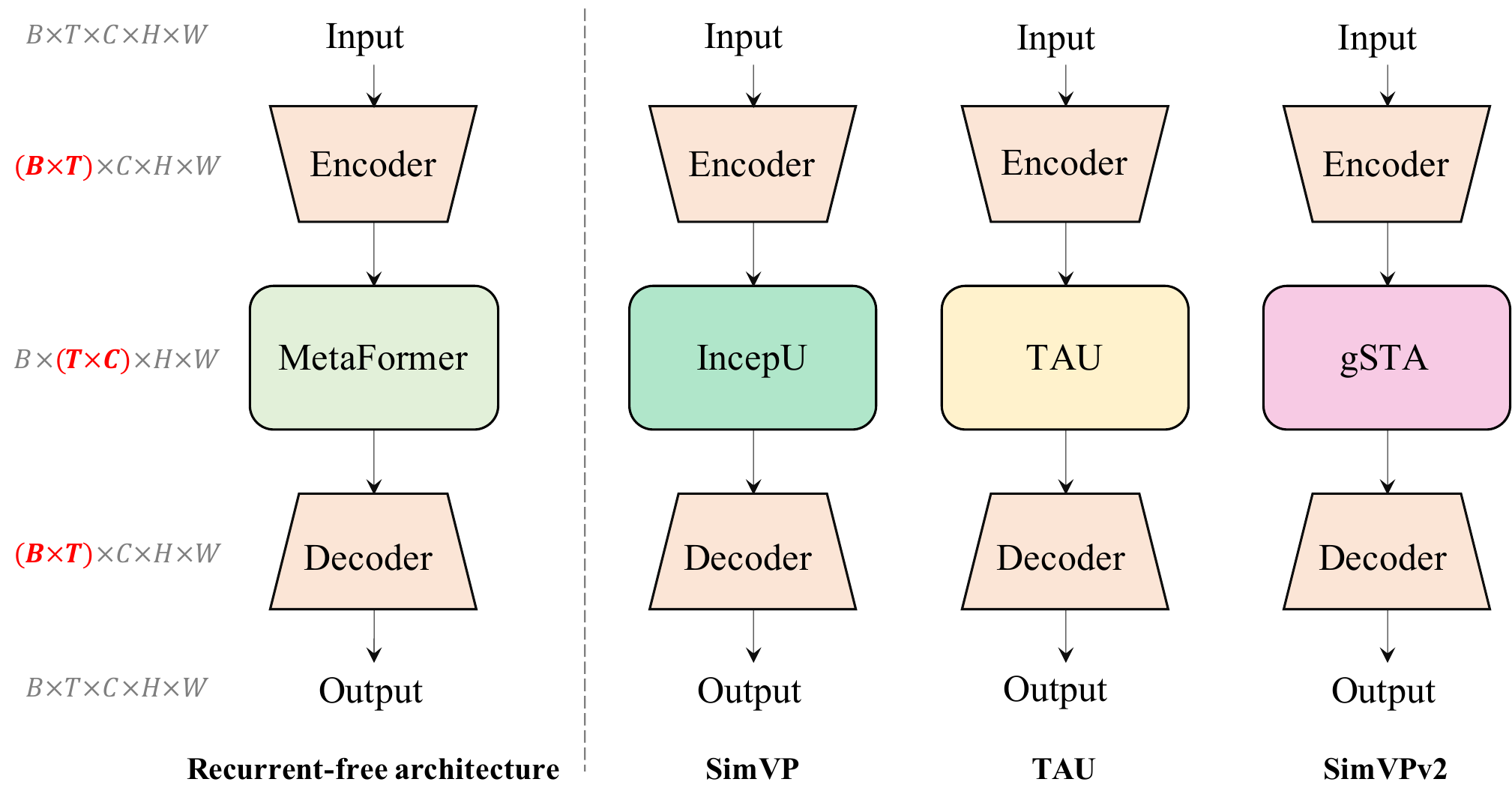}
  \vspace{-0.5em}
  \caption{The general architecture of recurrent-free models with three instances.}
  \label{fig:recurrent-free}
  \vspace{-2mm}
\end{figure}

Although less studied, recurrent-free spatio-temporal predictive learning models share a similar architecture, as illustrated in Figure~\ref{fig:recurrent-free}. The encoder comprises several 2D convolutional networks, which project high-dimensional input data into a low-dimensional latent space. When given a batch of input observed frames $\mathcal{B}\in\mathbb{R}^{B\times T\times C\times H\times W}$, the encoder focuses solely on intra-frame spatial correlations, ignoring temporal modeling. Subsequently, the middle temporal module stacks the low-dimensional representations along the temporal dimension to ascertain temporal dependencies. Finally, the decoder comprises several 2D convolutional upsampling networks, which reconstruct subsequent frames from the learned latent representations.

The encoder and decoder enable efficient temporal learning by modeling temporal dependencies in a low-dimensional latent space. The core component of recurrent-free models is the temporal module. Previous studies have proposed temporal modules such as IncepU~\cite{simvp}, TAU~\cite{tan2022temporal}, and gSTA~\cite{tan2022simvp} that have proved beneficial. However, we argue that the competence stems primarily from the general recurrent-free architecture instead of the specific temporal modules. Thus, we employ MetaFormers~\cite{yu2022metaformer} as the temporal module by changing the input channels from the original $C$ to inter-frame channels $T\times C$. By extending the recurrent-free architecture, we leverage the advantages of MetaFormers to enhance the recurrent-free model. In this work, we implement ViT~\cite{dosovitskiyimage}, Swin Transformer~\cite{liu2021swin}, Uniformer~\cite{li2022uniformer}, MLP-Mixer~\cite{tolstikhin2021mlp}, ConvMixer~\cite{trockman2022patches}, Poolformer~\cite{yu2022metaformer}, ConvNeXt~\cite{liu2022convnet}, VAN~\cite{guo2022visual}, HorNet~\cite{rao2022hornet}, and MogaNet~\cite{li2022efficient} for the MetaFormers-based recurrent-free model, substituting the intermediate temporal module in the original recurrent-free architecture.

\subsection{Supported Tasks}

We have curated five diverse tasks in our OpenSTL benchmark, which cover a wide range of scenarios from synthetic simulations to real-world situations at various scales. The tasks include synthetic moving object trajectories, real-world human motion capture, driving scenes, traffic flow, and weather forecasting. The datasets used in our benchmark range from synthetic to real-world, and from micro to macro scales. We have provided a summary of the dataset statistics in Table~\ref{tab:datasets}. 

\begin{table}[ht]
\small
\centering
\caption{The detailed dataset statistics of the supported tasks in OpenSTL.}
\setlength{\tabcolsep}{2.0mm}{
\begin{tabular}{cccccccc}
\toprule
Dataset  & Training size & Testing size & Channel & Height & Width & $T$ & $T'$ \\
\midrule
Moving MNIST variants & 10,000 & 10,000 & 1 / 3 & 64 & 64 & 10 & 10 \\
KTH Action    & 4,940 & 3,030 & 1 & 128 & 128 & 10 & 20/40 \\
Human3.6M  & 73,404 & 8,582 & 3 & 128 & 128 & 4 &  4 \\
Kitti\&Caltech & 3,160 & 3,095 & 3 & 128 & 160 & 10 & 1 \\
TaxiBJ & 20,461 & 500 & 2 & 32 & 32 & 4 & 4 \\
WeatherBench-S & 2,167 & 706 & 1 & 32/128 & 64/256 & 12 & 12 \\
WeatherBench-M & 54,019 & 2,883 & 4 & 32 & 64 & 4 & 4 \\
\bottomrule
\end{tabular}}
\label{tab:datasets}
\end{table}

\paragraph{Synthetic moving object trajectory prediction} \textit{Moving MNIST}~\cite{srivastava2015unsupervised} is one of the seminal benchmark datasets that has been extensively utilized. Each video sequence comprises two moving digits confined within a $64\times 64$ frame. Each digit was assigned a velocity whose direction was randomly chosen from a unit circle and whose magnitude was also arbitrarily selected from a fixed range. Apart from the original Moving MNIST dataset, we provide two variants with more complicated objects (\textit{Moving FashionMNIST}) that replace the digits with fashion objects and more complex scenes (\textit{Moving MNIST-CIFAR}) that employ images from the CIFAR-10 dataset~\cite{krizhevsky2009learning} as the background.
Moreover, we provide three settings of Moving MNIST for robustness evaluations, including missing frames, dynamic noise, and perceptual occlusions.

\paragraph{Human motion capture} Predicting human motion is challenging due to the complexity of human movements, which vary greatly among individuals and actions. We utilized the \textit{KTH} dataset~\cite{schuldt2004recognizing}, which includes six types of human actions: walking, jogging, running, boxing, hand waving, and hand clapping. We furnish two settings, predicting the next 20 and 40 frames respectively. \textit{Human3.6M}~\cite{ionescu2013human3} is an intricate human pose dataset containing high-resolution RGB videos. Analogous to preceding studies~\cite{phydnet,mim}, we predict the next four frames by the observed four frames.

\paragraph{Driving scene prediction} Predicting the future dynamics of driving scenarios is crucial for autonomous driving. Compared to other tasks, this undertaking involves non-stationary and diverse scenes. To address this issue, we follow the conventional approach~\cite{prednet} and train the model on the \textit{Kitti}~\cite{geiger2013vision} dataset. We then evaluate the performance on the \textit{Caltech Pedestrian}~\cite{dollarCVPR09peds} dataset. To ensure consistency, we center-cropped and downsized all frames to $128\times 160$ pixels. 
% By utilizing spatiotemporal predictive learning, the model is supposed to generate reasonable predictions even in challenging situations, such as when the camera-mounted vehicle is turning.

\paragraph{Traffic flow prediction} Forecasting the dynamics of crowds is crucial for traffic management and public safety. To evaluate spatio-temporal predictive learning approaches for traffic flow prediction, we use the \textit{TaxiBJ}~\cite{zhang2017deep} dataset. 
% This dataset includes GPS data from taxis and meteorological data in Beijing during four time periods: July 1, 2013 - October 30, 2013; March 1, 2014 - June 30, 2014; March 1, 2015 - June 30, 2015; and November 1, 2015 - April 10, 2016. 
This dataset includes GPS data from taxis and meteorological data in Beijing.
The dataset contains two types of crowd flows, representing inflow and outflow. The temporal interval is 30 minutes, and the spatial resolution is $32\times 32$. 

\paragraph{Weather forecasting} Global weather pattern prediction is an essential natural predicament. The WeatherBench~\cite{rasp2020weatherbench} dataset is a large-scale weather forecasting dataset encompassing various types of climatic factors. The raw data is re-grid to 5.625$^\circ$ resolution ($32\times 64$ grid points) and 1.40625$^\circ$ ($128\times 256$ grid points). We consider two setups: First, \textit{WeatherBench-S} is a single-variable setup in which each climatic factor is trained independently. The model is trained on data from 2010-2015, validated on data from 2016, and tested on data from 2017-2018, with a one-hour temporal interval. Second, \textit{WeatherBench-M} is a multi-variable setup that mimics real-world weather forecasting more closely. All climatic factors are trained simultaneously. The model is trained on data from 1979 to 2015, using the same validation and testing data as WeatherBench-S. The temporal interval is extended to six hours, capturing a broader range of temporal dependencies.

\subsection{Evaluation Metrics}

We evaluate the performance of supported models on the aforementioned tasks using various metrics in a thorough and rigorous manner. We use them for specific tasks according to their characteristics.

\noindent{\textbf{Error metrics}} We utilize the mean squared error (MSE) and mean absolute error (MAE) to evaluate the difference between the predicted results and the true targets. Root mean squared error (RMSE) is also used in weather forecasting as it is more common in this domain. 

\noindent{\textbf{Similarity metrics}} We utilize the structural similarity index measure (SSIM)~\cite{wang2004image} and peak signal-to-noise ratio (PSNR) to evaluate the similarity between the predicted results and the true targets. Such metrics are extensively used in image processing and computer vision. 

\noindent{\textbf{Perceptual metrics}} LPIPS~\cite{zhang2018unreasonable} is implemented to evaluate the perceptual difference between the predicted results and the true targets in the human visual system. LPIPS provides a perceptually-aligned evaluation for vision tasks. We utilize this metric in real-world video prediction tasks.

\noindent{\textbf{Computational metrics}} We utilize the number of parameters and the number of floating-point operations (FLOPs) to evaluate the computational complexity of the models. We also report the frames per second (FPS) on a single NVIDIA V100 GPU to evaluate the inference speed.

\subsection{Codebase Structure}

While existing open-sourced spatio-temporal predictive learning codebases are independent, OpenSTL provides a modular and extensible framework that adheres to the design principles of OpenMMLab~\cite{mmcv} and assimilates code elements from OpenMixup~\cite{li2022openmixup} and USB~\cite{wang2022usb}. OpenSTL excels in user-friendliness, organization, and comprehensiveness, surpassing the usability of existing open-source STL codebases. A detailed description of the codebase structure can be found in Appendix~\ref{app:codebase}.

% app:codebase

\section{Experiment and Analysis}

We conducted comprehensive experiments on the mentioned tasks to assess the performance of the supported methods in OpenSTL. Detailed analysis of the results is presented to gain insights into spatio-temporal predictive learning. Implementation details can be found in Appendix~\ref{app:implementation}.

\subsection{Synthetic Moving Object Trajectory Prediction}

We conduct experiments on the synthetic moving object trajectory prediction task, utilizing three datasets: Moving MNIST, Moving FashionMNIST, and Moving MNIST-CIFAR. The performance of the evaluated models on the Moving MNIST dataset is reported in Table~\ref{tab:mmnist}. The detailed results for the other two synthetic datasets are in Appendix~\ref{app:synthetic}. 

It can be observed that recurrent-based models yield varied results that do not consistently outperform recurrent-free models, while recurrent-based models always exhibit slower inference speeds than their recurrent-free counterparts. Although PredRNN, PredRNN++, MIM, and PredRNNv2 achieve lower MSE and MAE values compared to recurrent-free models, their FLOPs are nearly five times higher, and their FPS are approximately seven times slower than all recurrent-free models. Furthermore, there are minimal disparities in the performance of recurrent-free models as opposed to recurrent-based models, highlighting the robustness of the proposed general recurrent-free architecture. The remaining two synthetic datasets, consisting of more intricate moving objects (Moving FashionMNIST) and complex scenes (Moving MNIST-CIFAR), reinforce the experimental findings that recurrent-free models deliver comparable performance with significantly higher efficiency. In these toy datasets characterized by high frequency but low resolution, recurrent-based models excel in capturing temporal dependencies but are susceptible to high computational complexity. 
% Recurrent-free models strike a commendable balance between efficiency and performance.

\begin{table}[ht]
    \small
    \centering
    \setlength{\tabcolsep}{1.5mm}
    \caption{The performance on the Moving MNIST dataset.}
{\renewcommand\baselinestretch{1.5}\selectfont
  \resizebox{0.95\textwidth}{!}{
\begin{tabular}{ccccccccc}
\toprule
\multicolumn{2}{c}{Method}                                                    & Params (M) & FLOPs (G)  & FPS & MSE $\downarrow$ & MAE $\downarrow$ & SSIM $\uparrow$ & PSNR $\uparrow$ \\ \hline
\rowcolor[HTML]{FDF0E2} 
\cellcolor[HTML]{FDF0E2}                                   & ConvLSTM       & 15.0  & 56.8  & 113 & 29.80  & 90.64  & 0.9288 & 22.10 \\
\rowcolor[HTML]{FDF0E2} 
\cellcolor[HTML]{FDF0E2}                                   & PredNet          & 12.5  & 8.4 & 659 & 161.38 & 201.16 & 0.7783 & 14.67 \\
\rowcolor[HTML]{FDF0E2} 
\cellcolor[HTML]{FDF0E2}                                   & PredRNN          & 23.8  & 116.0 & 54  & 23.97  & 72.82  & 0.9462 & 23.28 \\
\rowcolor[HTML]{FDF0E2} 
\cellcolor[HTML]{FDF0E2}                                   & PredRNN++        & 38.6  & 171.7 & 38  & \textbf{22.06}  & \textbf{69.58}  & \textbf{0.9509} & \textbf{23.65} \\
\rowcolor[HTML]{FDF0E2} 
\cellcolor[HTML]{FDF0E2}                                   & MIM              & 38.0  & 179.2 & 37  & 22.55  & 69.97  & 0.9498 & 23.56 \\
\rowcolor[HTML]{FDF0E2} 
\cellcolor[HTML]{FDF0E2}                                   & E3D-LSTM         & 51.0  & 298.9 & 18  & 35.97  & 78.28  & 0.9320 & 21.11 \\
\rowcolor[HTML]{FDF0E2} 
\cellcolor[HTML]{FDF0E2}                                   & CrevNet          & 5.0   & 270.7 & 10  & 30.15  & 86.28  & 0.9350 & 22.15 \\
\rowcolor[HTML]{FDF0E2} 
\cellcolor[HTML]{FDF0E2}                                   & PhyDNet          & 3.1   & 15.3  & 182 & 28.19  & 78.64  & 0.9374 & 22.62 \\
\rowcolor[HTML]{FDF0E2} 
\cellcolor[HTML]{FDF0E2}                                   & MAU              & 4.5   & 17.8  & 201 & 26.86  & 78.22  & 0.9398 & 22.57 \\
\rowcolor[HTML]{FDF0E2} 
\cellcolor[HTML]{FDF0E2}                                   & PredRNNv2        & 23.9  & 116.6 & 52  & 24.13  & 73.73  & 0.9453 & 23.21 \\
\rowcolor[HTML]{FDF0E2} 
\multirow{-12}{*}{\cellcolor[HTML]{FDF0E2}Recurrent-based} & DMVFN            & 3.5   & 0.2   & 1145 & 123.67 & 179.96 & 0.8140 & 16.15 \\
\rowcolor[HTML]{E7ECE4} 
\cellcolor[HTML]{E7ECE4}                                   & SimVP            & 58.0  & 19.4  & 209 & 32.15  & 89.05  & 0.9268 & 21.84 \\
\rowcolor[HTML]{E7ECE4} 
\cellcolor[HTML]{E7ECE4}                                   & TAU              & 44.7  & 16.0  & 283 & 24.60  & 71.93  & 0.9454 & 23.19 \\
\rowcolor[HTML]{E7ECE4} 
\cellcolor[HTML]{E7ECE4}                                   & SimVPv2          & 46.8  & 16.5  & 282 & 26.69  & 77.19  & 0.9402 & 22.78 \\
\rowcolor[HTML]{E7ECE4} 
\cellcolor[HTML]{E7ECE4}                                   & ViT              & 46.1  & 16.9  & 290 & 35.15  & 95.87  & 0.9139 & 21.67 \\
\rowcolor[HTML]{E7ECE4} 
\cellcolor[HTML]{E7ECE4}                                   & Swin Transformer & 46.1  & 16.4  & 294 & 29.70  & 84.05  & 0.9331 & 22.22 \\
\rowcolor[HTML]{E7ECE4} 
\cellcolor[HTML]{E7ECE4}                                   & Uniformer        & 44.8  & 16.5  & 296 & 30.38  & 85.87  & 0.9308 & 22.13 \\
\rowcolor[HTML]{E7ECE4} 
\cellcolor[HTML]{E7ECE4}                                   & MLP-Mixer        & 38.2  & 14.7  & 334 & 29.52  & 83.36  & 0.9338 & 22.22 \\
\rowcolor[HTML]{E7ECE4} 
\cellcolor[HTML]{E7ECE4}                                   & ConvMixer        & 3.9   & 5.5   & 658 & 32.09  & 88.93  & 0.9259 & 21.93 \\
\rowcolor[HTML]{E7ECE4} 
\cellcolor[HTML]{E7ECE4}                                   & Poolformer       & 37.1  & 14.1  & 341 & 31.79  & 88.48  & 0.9271 & 22.03 \\
\rowcolor[HTML]{E7ECE4} 
\cellcolor[HTML]{E7ECE4}                                   & ConvNext         & 37.3  & 14.1  & 344 & 26.94  & 77.23  & 0.9397 & 22.74 \\
\rowcolor[HTML]{E7ECE4} 
\cellcolor[HTML]{E7ECE4}                                   & VAN              & 44.5  & 16.0  & 288 & 26.10  & 76.11  & 0.9417 & 22.89 \\
\rowcolor[HTML]{E7ECE4} 
\cellcolor[HTML]{E7ECE4}                                   & HorNet           & 45.7  & 16.3  & 287 & 29.64  & 83.26  & 0.9331 & 22.26 \\
\rowcolor[HTML]{E7ECE4} 
\multirow{-13}{*}{\cellcolor[HTML]{E7ECE4}Recurrent-free}  & MogaNet & 46.8  & 16.5  & 255 & 25.57  & 75.19  & 0.9429 & 22.99 \\
\bottomrule
\end{tabular}}\par}
\label{tab:mmnist}
\end{table}

\subsection{Real-world Video Prediction}

We perform experiments on real-world video predictions, specifically focusing on human motion capturing using the KTH and Human3.6M datasets, as well as driving scene prediction using the Kitti\&Caltech dataset. Due to space constraints, we present the results for the Kitti\&Caltech dataset in Table~\ref{tab:caltech}, while the detailed results for the other datasets can be found in Appendix~\ref{app:realworld}. We observed that as the resolution increases, the computational complexity of recurrent-based models dramatically increases. In contrast, recurrent-free models achieve a commendable balance between efficiency and performance. Notably, although some recurrent-based models achieve lower MSE and MAE values, their FLOPs are nearly 20 times higher compared to their recurrent-free counterparts. This highlights the efficiency advantage of recurrent-free models, especially in high-resolution scenarios.

\begin{table}[ht]
    \small
    \centering
    \setlength{\tabcolsep}{1.5mm}
    \caption{The performance on the Kitti\&Caltech dataset.}
{\renewcommand\baselinestretch{1.5}\selectfont
  \resizebox{0.95\textwidth}{!}{
\begin{tabular}{cccccccccc}
\toprule
\multicolumn{2}{c}{Method}                                    & Params (M) & FLOPs (G)  & FPS & MSE $\downarrow$ & MAE $\downarrow$ & SSIM $\uparrow$ & PSNR $\uparrow$ & LPIPS $\downarrow$\\ \hline
\rowcolor[HTML]{FDF0E2} 
\cellcolor[HTML]{FDF0E2}                                   & ConvLSTM       & 15.0  & 595.0  & 33 & 139.6  & 1583.3 & 0.9345 & 27.46 & 8.58 \\
\rowcolor[HTML]{FDF0E2} 
\cellcolor[HTML]{FDF0E2}                                   & PredNet       & 12.5  & 42.8 & 94 & 159.8 & 1568.9 & 0.9286 & 27.21 & 11.29 \\
\rowcolor[HTML]{FDF0E2} 
\cellcolor[HTML]{FDF0E2}                                   & PredRNN          & 23.7  & 1216.0 & 17 & 130.4 & 1525.5 & 0.9374 & 27.81 & 7.40 \\
\rowcolor[HTML]{FDF0E2} 
\cellcolor[HTML]{FDF0E2}                                   & PredRNN++        & 38.5 & 1803.0 & 12  & \textbf{125.5}  & \textbf{1453.2}  & 0.9433 & 28.02 & 13.21 \\
\rowcolor[HTML]{FDF0E2} 
\cellcolor[HTML]{FDF0E2}                                   & MIM       & 49.2  & 1858.0 & 39 & 125.1  & 1464.0  & 0.9409 & \textbf{28.10} & 6.35 \\
\rowcolor[HTML]{FDF0E2} 
\cellcolor[HTML]{FDF0E2}                                   & E3D-LSTM         & 54.9  & 1004.0 & 10 & 200.6  & 1946.2  & 0.9047 & 25.45 & 12.60 \\
\rowcolor[HTML]{FDF0E2} 
\cellcolor[HTML]{FDF0E2}                                   & PhyDNet          & 3.10 & 40.4 & 117 & 312.2 & 2754.8 & 0.8615 & 23.26 & 32.19 \\
\rowcolor[HTML]{FDF0E2} 
\cellcolor[HTML]{FDF0E2}                                   & MAU              & 24.3   & 172.0 & 16 & 177.8 & 1800.4 & 0.9176 & 26.14 & 9.67 \\
\rowcolor[HTML]{FDF0E2} 
\cellcolor[HTML]{FDF0E2}                                   & PredRNNv2        & 23.8  & 1223.0 & 16 & 147.8  & 1610.5 & 0.9330 & 27.12 & 8.92 \\
\rowcolor[HTML]{FDF0E2} 
\multirow{-12}{*}{\cellcolor[HTML]{FDF0E2}Recurrent-based} & DMVFN            & 3.6 & 1.2 & 557 & 183.9 & 1531.1 & 0.9314 & 26.78 & \textbf{4.94} \\
\rowcolor[HTML]{E7ECE4} 
\cellcolor[HTML]{E7ECE4}                                   & SimVP            & 8.6 & 60.6 & 57 & 160.2 & 1690.8 & 0.9338 & 26.81 & 6.76 \\
\rowcolor[HTML]{E7ECE4} 
\cellcolor[HTML]{E7ECE4}                                   & TAU              &  15.0 & 92.5 & 55 & 131.1 & 1507.8 & 0.9456 & 27.83 & 5.49 \\
\rowcolor[HTML]{E7ECE4} 
\cellcolor[HTML]{E7ECE4}                                   & SimVPv2          & 15.6 & 96.3  & 40 & 129.7 & 1507.7 & 0.9454 & 27.89 & 5.57 \\
\rowcolor[HTML]{E7ECE4} 
\cellcolor[HTML]{E7ECE4}                                   & ViT              & 12.7 & 155.0 & 25 & 146.4  & 1615.8  & 0.9379 & 27.43 & 6.66 \\
\rowcolor[HTML]{E7ECE4} 
\cellcolor[HTML]{E7ECE4}                                   & Swin Transformer & 15.3 & 95.2  & 49 & 155.2  & 1588.9  & 0.9299 & 27.25 & 8.11 \\
\rowcolor[HTML]{E7ECE4} 
\cellcolor[HTML]{E7ECE4}                                   & Uniformer        & 11.8  & 104.0  & 28 & 135.9  & 1534.2 & 0.9393 & 27.66 & 6.87 \\
\rowcolor[HTML]{E7ECE4} 
\cellcolor[HTML]{E7ECE4}                                   & MLP-Mixer        & 22.2  & 83.5  & 60 & 207.9  & 1835.9  & 0.9133 & 26.29 & 7.75 \\
\rowcolor[HTML]{E7ECE4} 
\cellcolor[HTML]{E7ECE4}                                   & ConvMixer        & 1.5   & 23.1   & 129 & 174.7  & 1854.3 & 0.9232 & 26.23 & 7.76 \\
\rowcolor[HTML]{E7ECE4} 
\cellcolor[HTML]{E7ECE4}                                   & Poolformer       & 12.4  & 79.8  & 51 & 153.4  & 1613.5 & 0.9334 & 27.38 & 7.00 \\
\rowcolor[HTML]{E7ECE4} 
\cellcolor[HTML]{E7ECE4}                                   & ConvNext         & 12.5  & 80.2  & 54 & 146.8  & 1630.0 & 0.9336 & 27.19 & 6.99 \\
\rowcolor[HTML]{E7ECE4} 
\cellcolor[HTML]{E7ECE4}                                   & VAN              & 14.9  & 92.5  & 41 & 127.5  & 1476.5 & 0.9462 & 27.98 & 5.50 \\
\rowcolor[HTML]{E7ECE4} 
\cellcolor[HTML]{E7ECE4}                                   & HorNet           & 15.3  & 94.4  & 43 & 152.8  & 1637.9 & 0.9365 & 27.09 & 6.00 \\
\rowcolor[HTML]{E7ECE4} 
\multirow{-13}{*}{\cellcolor[HTML]{E7ECE4}Recurrent-free}  & MogaNet & 15.6  & 96.2  & 36 & 131.4 & 1512.1 & \textbf{0.9442} & 27.79 & 5.39 \\
\bottomrule
\end{tabular}}\par}
\label{tab:caltech}
\end{table}

\subsection{Traffic and Weather Forecasting}

Traffic flow prediction and weather forecasting are two critical tasks that have significant implications for public safety and scientific research. While these tasks operate at a macro level, they exhibit lower frequencies compared to the tasks mentioned above, and the states along the timeline tend to be more stable. Capturing subtle changes in such tasks poses a significant challenge. In order to assess the performance of the supported models in OpenSTL, we conduct experiments on the TaxiBJ and WeatherBench datasets. It is worth noting that weather forecasting encompasses various settings, and we provide detailed results of them in Appendix~\ref{app:trafficweather}.

\begin{figure}[t]
  \centering
  \vspace{-2.0em}
  \includegraphics[width=1.0\textwidth]{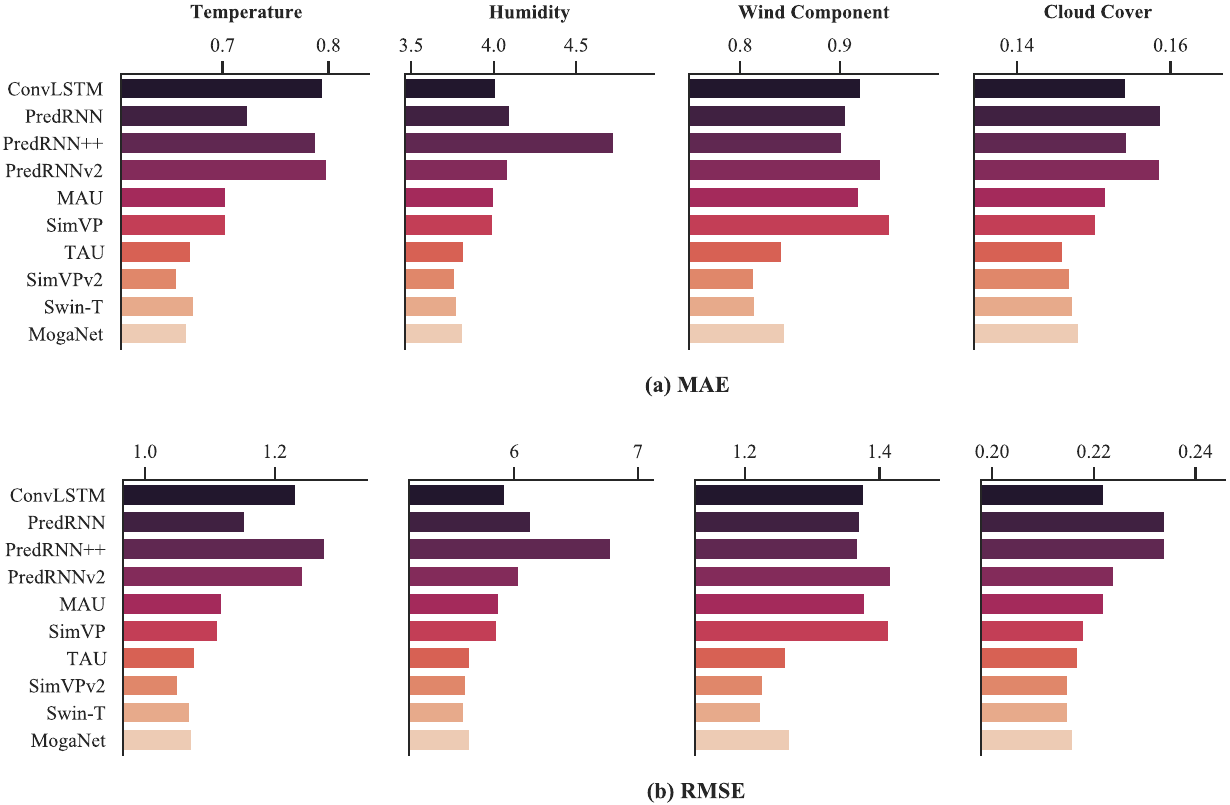}
  \vspace{-1.0em}
  \caption{The (a) MAE and (b) RMSE metrics of the representative approaches on the four weather forecasting tasks in WeatherBench.}
  \label{fig:weatherbench}
  \vspace{-0.5em}
\end{figure}

Here, we present a comparison of the MAE and RMSE metrics for representative approaches in single-variable weather factor forecasting at low resolution. Figure~\ref{fig:weatherbench} displays the results for four climatic factors, i.e., temperature, humidity, wind component, and cloud cover. Notably, recurrent-free models consistently outperform recurrent-based models across all weather factors, indicating their potential to apply spatio-temporal predictive learning to macro-scale tasks instead of relying solely on recurrent-based models. These findings underscore the promising nature of recurrent-free models and suggest that they can be a viable alternative to the prevailing recurrent-based models in the context of weather forecasting. Furthermore, in the Appendix, we provide additional insights into high-resolution and multi-variable weather forecasting, where similar trends are observed.

\subsection{Robustness Analysis}

To further understand the differences in robustness between the recurrent-based and recurrent-free spatiotemporal predictive learning methods, we constructed three experimental setups: (i) Moving MNIST - Missing, which deals with input frames with missing frames, where we set the probability of random missing frame to 20\%; (ii) Moving MNIST - Dynamic, where we added random Gaussian noise to the speed of each digit, making their movement speeds irregular; (iii) Moving MNIST - Perceptual, where we randomly occluded the input frames, using a black 24$\times$24 patch for occlusion. We choose three representative recurrent-based and three recurrent-free methods for evaluation. The experimental results for these three setups are presented in Table~\ref{tab:mmnist_missing}, \ref{tab:mmnist_dynamic}, \ref{tab:mmnist_perceptual}, respectively.

It can be observed that the recurrent-free methods exhibit remarkable robustness under both the missing and perceptual noise scenarios. Even when compared to situations without noise, there is little performance degradation due to their focus on global information. Conversely, recurrent-based methods encounter substantial performance drops. They overly focus on the relationships individual frames can inadvertently lead to overfitting. In the case of dynamic noise, all methods faced significant performance setbacks, because the speed of the digits became irregular and harder to predict.

\begin{table}[b]
  \small
  \centering
  \setlength{\tabcolsep}{1.5mm}
  \caption{The performance on the Moving MNIST - Missing dataset.}
  {\renewcommand\baselinestretch{1.5}\selectfont
    \resizebox{0.92\textwidth}{!}{
  \begin{tabular}{ccccccccc}
  \toprule
  \multicolumn{2}{c}{Method}                                                    & Params (M) & FLOPs (G)  & FPS & MSE $\downarrow$ & MAE $\downarrow$ & SSIM $\uparrow$ & PSNR $\uparrow$ \\ \hline
  \rowcolor[HTML]{FDF0E2} 
  \cellcolor[HTML]{FDF0E2}                                   & ConvLSTM       & 15.0  & 56.8  & 113 & 32.73  & 96.95  & 0.9201 & 21.65 \\
  \rowcolor[HTML]{FDF0E2} 
  \cellcolor[HTML]{FDF0E2}                                   & PredRNN          & 23.8  & 116.0 & 54  & 46.05  & 117.21  & 0.8800 & 20.35 \\
  \rowcolor[HTML]{FDF0E2} 
  \multirow{-3}{*}{\cellcolor[HTML]{FDF0E2}Recurrent-based}                                  & PredRNN++        & 38.6  & 171.7 & 38  & 53.89  & 118.45  & 0.8907  & 19.71 \\
  \rowcolor[HTML]{E7ECE4} 
  \cellcolor[HTML]{E7ECE4}                                   & SimVP            & 58.0  & 19.4  & 209 & 34.92 & 95.23 & 0.9194 & 21.44 \\
  \rowcolor[HTML]{E7ECE4} 
  \cellcolor[HTML]{E7ECE4}                                   & TAU              & 44.7  & 16.0  & 283 & \textbf{26.77} & \textbf{77.50} & \textbf{0.9400} & \textbf{22.74} \\
  \rowcolor[HTML]{E7ECE4} 
  \multirow{-3}{*}{\cellcolor[HTML]{E7ECE4}Recurrent-free} & SimVPv2          & 46.8  & 16.5  & 282 & 28.63 & 81.79 & 0.9352 & 22.39 \\
  \bottomrule
  \end{tabular}}\par}
  \label{tab:mmnist_missing}
\end{table}

\begin{table}[t]
  \small
  \vspace{-1.0em}
  \centering
  \setlength{\tabcolsep}{1.5mm}
  \caption{The performance on the Moving MNIST - Dynamic dataset.}
  {\renewcommand\baselinestretch{1.5}\selectfont
    \resizebox{0.92\textwidth}{!}{
  \begin{tabular}{ccccccccc}
  \toprule
  \multicolumn{2}{c}{Method}                                                    & Params (M) & FLOPs (G)  & FPS & MSE $\downarrow$ & MAE $\downarrow$ & SSIM $\uparrow$ & PSNR $\uparrow$ \\ \hline
  \rowcolor[HTML]{FDF0E2} 
  \cellcolor[HTML]{FDF0E2}                                   & ConvLSTM       & 15.0  & 56.8  & 113 & 49.03 & 135.49 & 0.8683 & 19.73 \\
  \rowcolor[HTML]{FDF0E2} 
  \cellcolor[HTML]{FDF0E2}                                   & PredRNN          & 23.8  & 116.0 & 54  & 59.18 & 157.47  & 0.8220 & 19.09 \\
  \rowcolor[HTML]{FDF0E2} 
  \multirow{-3}{*}{\cellcolor[HTML]{FDF0E2}Recurrent-based}                                  & PredRNN++        & 38.6  & 171.7 & 38  & \textbf{40.85} & \textbf{109.32} & \textbf{0.9030} & \textbf{20.65} \\
  \rowcolor[HTML]{E7ECE4} 
  \cellcolor[HTML]{E7ECE4}                                   & SimVP            & 58.0  & 19.4  & 209 & 48.41 & 130.83 & 0.8725 & 19.91 \\
  \rowcolor[HTML]{E7ECE4} 
  \cellcolor[HTML]{E7ECE4}                                   & TAU              & 44.7  & 16.0  & 283 & 43.37 & 121.31 & 0.8853 & 20.41 \\
  \rowcolor[HTML]{E7ECE4} 
  \multirow{-3}{*}{\cellcolor[HTML]{E7ECE4}Recurrent-free} & SimVPv2          & 46.8  & 16.5  & 282 & 44.74 & 123.70 & 0.8823 & 20.28 \\
  \bottomrule
  \end{tabular}}\par}
  \label{tab:mmnist_dynamic}
  \vspace{-0.5em}
\end{table}

\begin{table}[t]
  \small
  \centering
  \setlength{\tabcolsep}{1.5mm}
  \caption{The performance on the Moving MNIST - Perceptual dataset.}
  {\renewcommand\baselinestretch{1.5}\selectfont
    \resizebox{0.92\textwidth}{!}{
  \begin{tabular}{ccccccccc}
  \toprule
  \multicolumn{2}{c}{Method}                                                    & Params (M) & FLOPs (G)  & FPS & MSE $\downarrow$ & MAE $\downarrow$ & SSIM $\uparrow$ & PSNR $\uparrow$ \\ \hline
  \rowcolor[HTML]{FDF0E2} 
  \cellcolor[HTML]{FDF0E2}                                   & ConvLSTM       & 15.0  & 56.8  & 113 & 31.34 & 95.39 & 0.9227 & 21.85 \\
  \rowcolor[HTML]{FDF0E2} 
  \cellcolor[HTML]{FDF0E2}                                   & PredRNN          & 23.8  & 116.0 & 54  & 46.04 & 122.40 & 0.8792 & 20.28 \\
  \rowcolor[HTML]{FDF0E2} 
  \multirow{-3}{*}{\cellcolor[HTML]{FDF0E2}Recurrent-based}                                  & PredRNN++        & 38.6  & 171.7 & 38  & 51.76  & 127.12 & 0.8722 & 19.85 \\
  \rowcolor[HTML]{E7ECE4} 
  \cellcolor[HTML]{E7ECE4}                                   & SimVP            & 58.0  & 19.4  & 209 & 34.73 & 95.23 & 0.9196 & 21.44 \\
  \rowcolor[HTML]{E7ECE4} 
  \cellcolor[HTML]{E7ECE4}                                   & TAU              & 44.7  & 16.0  & 283 & \textbf{26.87} & \textbf{78.08} & \textbf{0.9393} & \textbf{22.69} \\
  \rowcolor[HTML]{E7ECE4} 
  \multirow{-3}{*}{\cellcolor[HTML]{E7ECE4}Recurrent-free} & SimVPv2 & 46.8  & 16.5  & 282 & 28.83 & 82.65 & 0.9343 & 22.36 \\
  \bottomrule
  \end{tabular}}\par}
  \label{tab:mmnist_perceptual}
  \vspace{-0.5em}
\end{table}

\section{Conclusion and Discussion}

This paper introduces OpenSTL, a comprehensive benchmark for spatio-temporal predictive learning with a diverse set of 14 representative methods and 24 models, addressing a wide range of challenging tasks. OpenSTL categorizes existing approaches into recurrent-based and recurrent-free models. To unlock the potential of recurrent-free models, we propose a general recurrent-free architecture and introduce MetaFormers for temporal modeling. Extensive experiments are conducted to systematically evaluate the performance of the supported models across various tasks. In synthetic datasets, recurrent-based models excel at capturing temporal dependencies, while recurrent-free models achieve comparable performance with significantly higher efficiency. In real-world video prediction tasks, recurrent-free models strike a commendable balance between efficiency and performance. Additionally, recurrent-free models demonstrate significant superiority over their counterparts in weather forecasting, highlighting their potential for scientific applications at a macro-scale level. 

Moreover, we observed that \textit{recurrent architectures are beneficial in capturing temporal dependencies, but they are not always necessary, especially for computationally expensive tasks}. Recurrent-free models can be a viable alternative that provides a good balance between efficiency and performance. The effectiveness of recurrent-based models in capturing high-frequency spatio-temporal dependencies can be attributed to their sequential tracking of frame-by-frame changes, providing a local temporal inductive bias. On the other hand, recurrent-free models combine multiple frames together, exhibiting a global temporal inductive bias that is suitable for low-frequency spatio-temporal dependencies. We hope that our work provides valuable insights and serves as a reference for future research. 

While our primary focus lies in general spatio-temporal predictive learning, there are still several open problems that require further investigation. One particular challenge is finding ways to effectively leverage the strengths of both recurrent-based and recurrent-free models to enhance the modeling of spatial-temporal dependencies. While there is a correspondence between the spatial encoding and temporal modeling in MetaVP and the token mixing and channel mixing in MetaFormer, it raises the question of whether we can improve recurrent-free models by extending the existing MetaFormers.
% While our primary focus lies in general spatiotemporal predictive learning, there are still several open problems in this field that require further investigation. For example, it remains to be explored whether recurrent-free models can provide advantages in higher-resolution real-world video prediction scenarios. Additionally, although we have considered several important climatic factors in weather forecasting, it is necessary to determine whether the conclusions hold for more intricate weather phenomena. We view these questions as promising directions for future research, offering exciting opportunities for further exploration and advancements in the field.

\begin{ack}
This work was supported by the National Key R\&D Program of China (2022ZD0115100), the National Natural Science Foundation of China (U21A20427), the Competitive Research Fund (WU2022A009) from the Westlake Center for Synthetic Biology and Integrated Bioengineering.
% Use unnumbered first level headings for the acknowledgments. All acknowledgments
% go at the end of the paper before the list of references. Moreover, you are required to declare
% funding (financial activities supporting the submitted work) and competing interests (related financial activities outside the submitted work).
% More information about this disclosure can be found at: \url{https://neurips.cc/Conferences/2023/PaperInformation/FundingDisclosure}.
% You can use the \texttt{ack} environment provided in the style file. As opposed to the main NeurIPS track, acknowledgements do not need to be hidden.
\end{ack}

\bibliography{ref}
\bibliographystyle{abbrv}

\newpage

\appendix

\section{Installation and Data Preparation} 
\label{app:datasets}

\subsection{Installation}

In our GitHub (\href{github.com/chengtan9907/OpenSTL}{github.com/chengtan9907/OpenSTL}), we have provided a conda environment setup file for OpenSTL. Users can easily reproduce the environment by executing the following commands:
\begin{lstlisting}[language=Python, 
  frame=single,
  basicstyle=\small\ttfamily,
  keywordstyle=\color{blue},
  stringstyle=\color{red},
  commentstyle=\color{black}]
  git clone https://github.com/chengtan9907/OpenSTL
  cd OpenSTL
  conda env create -f environment.yml
  conda activate OpenSTL
  python setup.py develop  \# or "pip install -e ."
\end{lstlisting}

By following the instructions above, OpenSTL will be installed in development mode, allowing any local code modifications to take effect. Alternatively, users can install it as a PyPi package using \texttt{pip install .}, but remember to reinstall it to apply any local modifications.

\subsection{Data Preparation}

It is recommended to symlink the dataset root (assuming \texttt{\$USER\_DATA\_ROOT}) to \texttt{\$OpenSTL/data}. If the folder structure of the user is different, the user needs to change the corresponding paths in config files. We provide tools to download and preprocess the datasets in \texttt{OpenSTL/tool/prepare\_data}.

\section{Codebase Overview}
\label{app:codebase}

\begin{figure}[ht]
  \centering
  \includegraphics[width=0.98\textwidth]{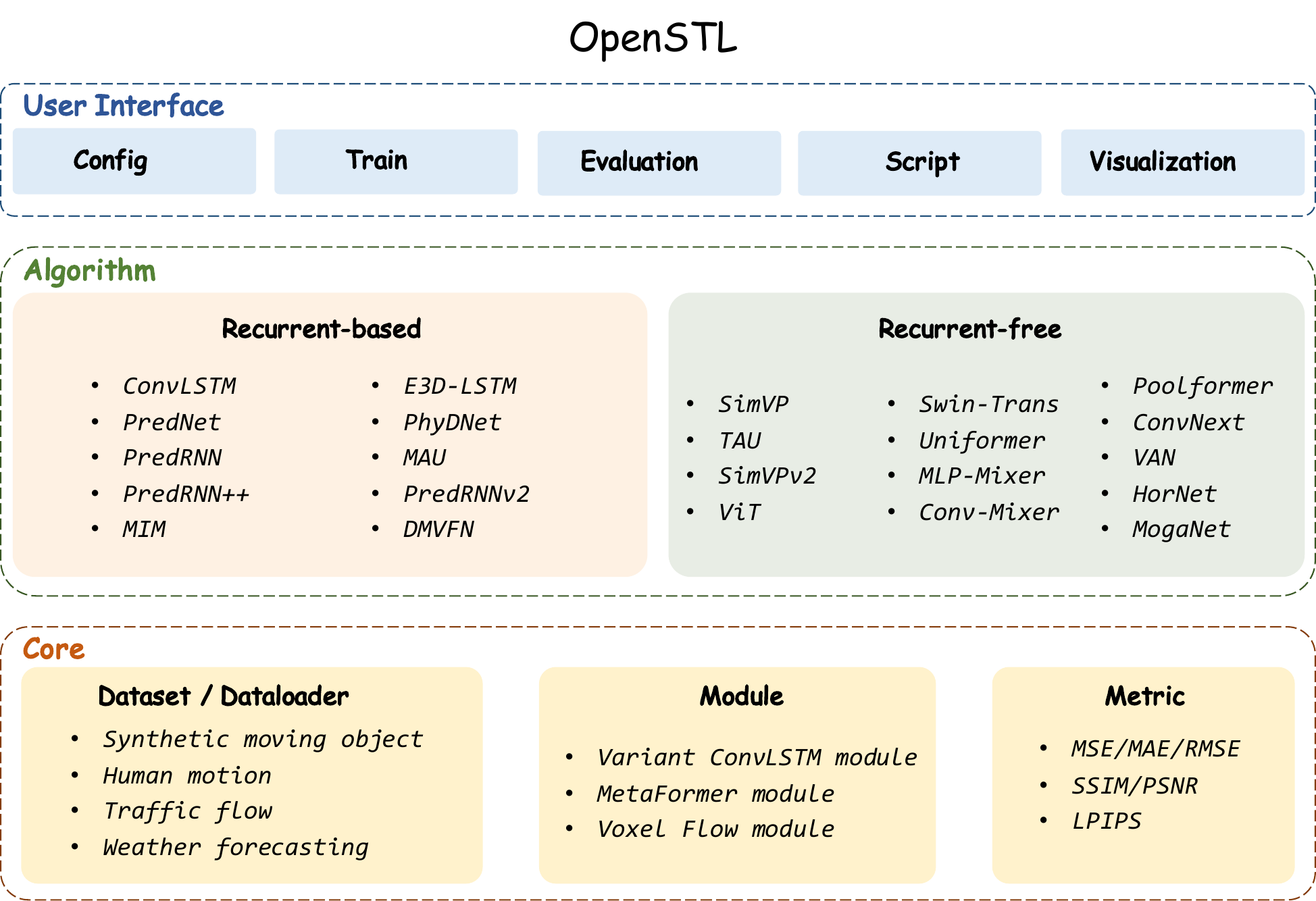}
  \caption{The graphical overview of OpenSTL.}
  \label{fig:codebase}
\end{figure}

In this section, we present a comprehensive overview of the codebase structure of OpenSTL. The codebase is organized into three abstracted layers, namely the core layer, algorithm layer, and user interface layer, arranged from the bottom to the top, as illustrated in Figure~\ref{fig:codebase}.

\paragraph{Core Layer} The core layer comprises essential components of OpenSTL, such as dataloaders for supported datasets, basic modules for supported models, and metrics for evaluation. The dataloaders offer a unified interface for data loading and preprocessing. The modules consist of foundational unit implementations of supported models. The metrics provide a unified interface for evaluation purposes. The core layer establishes a foundation for the upper layers to ensure flexibility in usage.

\paragraph{Algorithm Layer} The algorithm layer encompasses the implementations of the supported models, which are organized into two distinct categories: recurrent-based and recurrent-free models. These implementations are developed using the PyTorch framework and closely adhere to the methodologies described in the original research papers and their official open-sourced code. The algorithm layer ensures the compatibility, reliability, and reproducibility of the supported algorithms by abstracting common components and avoiding code duplication, enabling the easy and flexible implementation of customized algorithms. Moreover, the algorithm layer provides a unified interface that facilitates seamless operations such as model training, evaluation, and testing. By offering a consistent interface, the algorithm layer enhances usability and promotes ease of experimentation with the models.

\paragraph{User Interface Layer} The user interface layer comprises configurations, training, evaluation, and scripts that facilitate the basic usage of OpenSTL. We offer convenient tools for generating visualizations. The user interface layer is designed to be user-friendly and intuitive, enabling users to easily train, evaluate, and test the supported algorithms. By offering detailed parameter settings in the configurations, the user interface layer provides a unified interface that enables users to reproduce the results presented in this paper, without requiring any additional efforts.
  
\section{Implementation Details}
\label{app:implementation}

Table~\ref{tab:implementation} describes the hyper-parameters employed in the supported models across multiple datasets, namely MMNIST, KITTI, KTH, Human, TaxiBJ, Weather-S, and Weather-M. For each dataset, the hyperparameters include $T$, $T'$, $\mathrm{hid}_S$, $\mathrm{hid}_T$, $N_S$, $N_T$, epoch, optimizer, drop path, and learning rate. $T$ and $T'$ have the same values for the MMNIST, Human, TaxiBJ, and Weather-M datasets, but differ for KITTI and KTH. The specific values of $T$ and $T'$ are depended on the dataset. The learning rate and drop path are chosen from a set of values, and the best result for each experiment is reported.

The parameters $\mathrm{hid}_S$ and $\mathrm{hid}_T$ correspond to the size of the hidden layers in the spatial encoder/decoder and the temporal module of the model, respectively. While these parameters exhibit minor variations across datasets, their values largely maintain consistency, underscoring the standardized model structure across diverse datasets. $N_S$ and $N_T$ denote the number of blocks in the spatial encoder/decoder and the temporal module, respectively. These four hyper-parameters are from recurrent-free models, we provide the detailed hyper-parameters of recurrent-based models in GitHub for theirs are various. Please refer to the link \href{https://github.com/chengtan9907/OpenSTL/tree/master/configs}{OpenSTL/configs} for more details.

\begin{table}[ht]
\centering
\caption{Hyper-parameters of the supported models.}
{\renewcommand\baselinestretch{1.1}\selectfont
  \resizebox{\textwidth}{!}{
\begin{tabular}{ccccccccc}
\toprule
Dataset & MMNIST & KITTI & KTH & Human & TaxiBJ & Weather-S & Weather-M \\
\midrule
$T$ & 10 & 10 & 10 & 4 & 4 & 12 & 4 \\
$T'$ & 10 & 1 & 20 & 4 & 4 & 12 & 4 \\
$\mathrm{hid}_S$ & 64 & 64 & 64 & 64 & 32 & 32 & 32 \\
$\mathrm{hid}_T$ & 512 & 256 & 256 & 512 & 256 & 256 & 256 \\
$N_S$            & 4   & 2   & 2   & 4   & 2   & 2   & 2 \\
$N_T$            & 8   & 6   & 6   & 6   & 8   & 8   & 8 \\
epoch            & 200 & 100 & 100 & 50  & 50  & 50  & 50 \\
optimizer & \multicolumn{7}{c}{Adam} \\
drop path& \multicolumn{7}{c}{\multirow{1}{*}{$\{0.0, 0.1, 0.2\}$}} \\
learning rate & \multicolumn{7}{c}{\multirow{1}{*}{\{$1e^{-2}, 5e^{-3}, 1e^{-3}, 5e^{-4}, 1e^{-4}$\}}} \\
\bottomrule
\end{tabular}}\par}
\label{tab:implementation}
\end{table}
  
\section{Detailed Experimental Results}

\subsection{Synthetic Moving Object Trajectory Prediction}
\label{app:synthetic}

\begin{figure}[ht]
  \centering
  \includegraphics[width=1.0\textwidth]{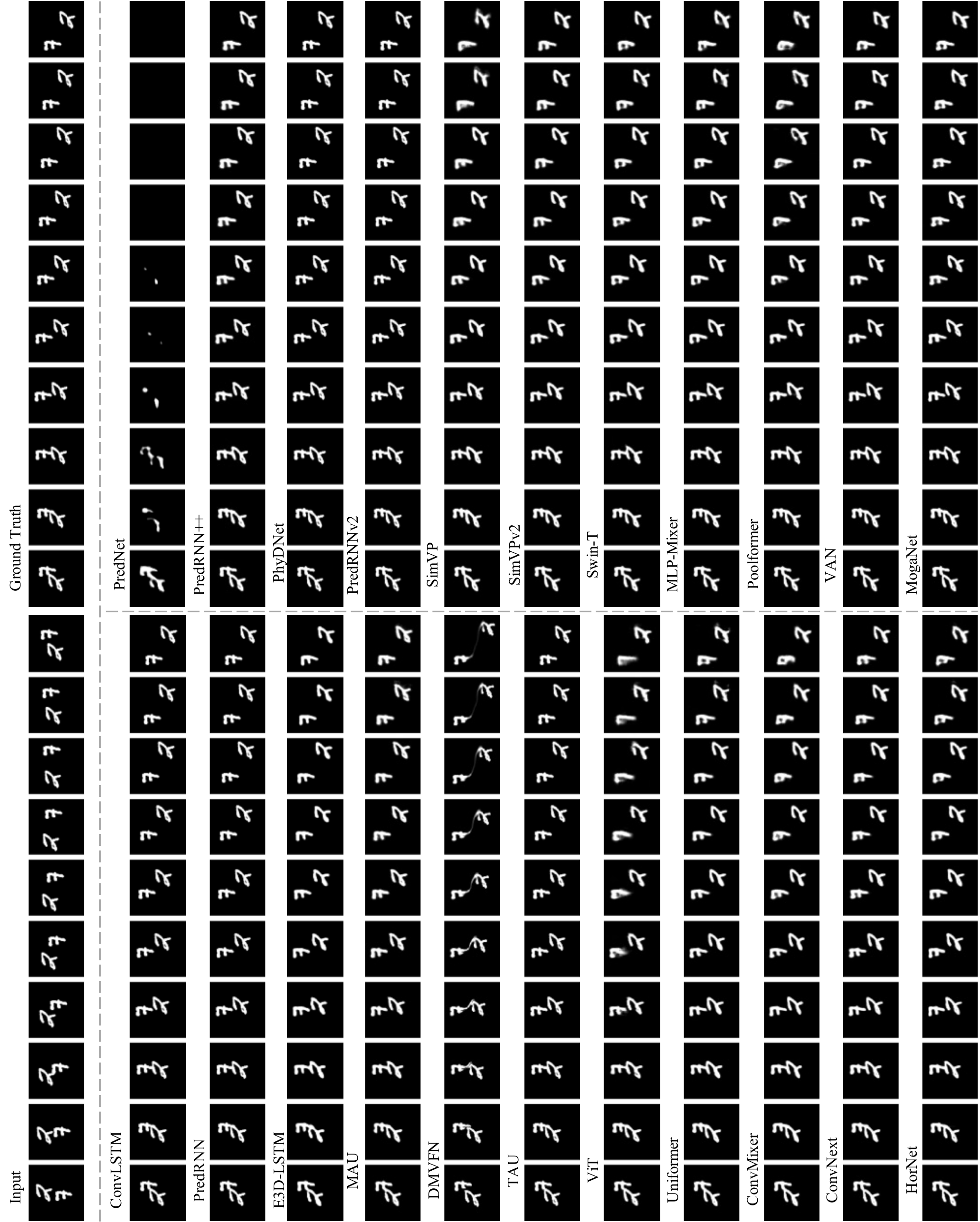}
  \caption{The qualitative visualization on Moving MNIST. For the convenience of formatting, we arrange the frames vertically from bottom to top.}
  \label{fig:visual_mmnist}
  \vspace{-2mm}
\end{figure}

\noindent{\textbf{Moving MNIST}} In addition to the quantitative results provided in the main text, we also provide a visualization example for qualitative assessment, as shown in Figure~\ref{fig:visual_mmnist}. For the convenience of formatting, we arrange the frames vertically from bottom to top. It can be observed that the majority of recurrent-based models produce high-quality predicted results, except for PredNet and DMVFN. Recurrent-free models achieve comparable results but exhibit blurriness in the last few frames. This phenomenon suggests that recurrent-based models excel at capturing temporal dependencies.

\begin{figure}[ht]
  \centering
  \includegraphics[width=1.0\textwidth]{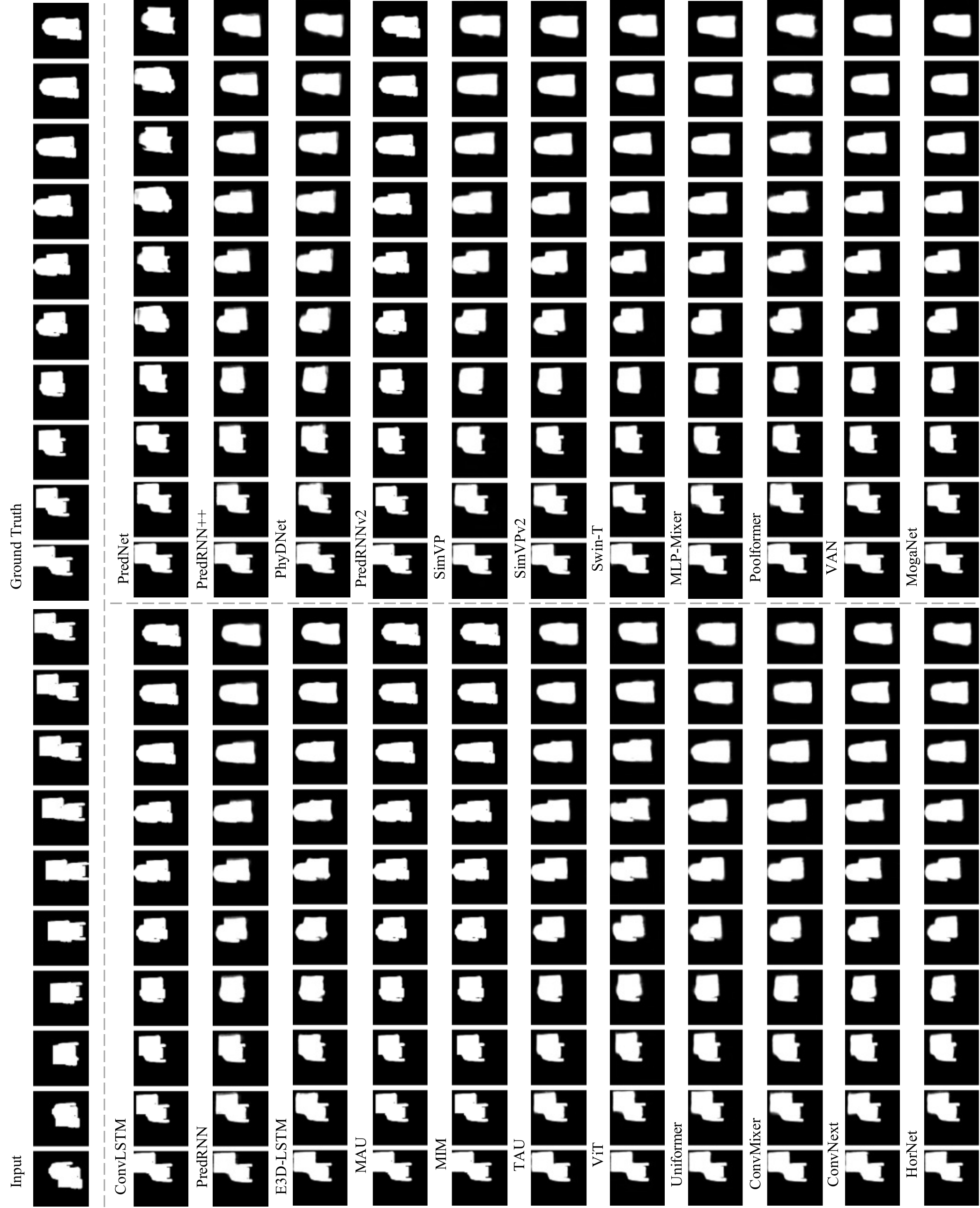}
  \caption{The qualitative visualization on Moving Fashion MNIST. For the convenience of formatting, we arrange the frames vertically from bottom to top.}
  \label{fig:visual_mfmnist}
\end{figure}

\noindent{\textbf{Moving FashionMNIST}} We show the quantitative results and qualitative visualization examples in Table~\ref{tab:fmmnist} and Figure~\ref{fig:visual_mfmnist}, respectively. The results are consistent with those of Moving MNIST, where recurrent-based models perform well in long-range temporal modeling.

\begin{table}[ht]
\vspace{-1mm}
\centering
\caption{The performance on the Moving FashionMNIST dataset.}
{\renewcommand\baselinestretch{1.18}\selectfont
  \resizebox{\textwidth}{!}{
\begin{tabular}{ccccccccc}
\toprule
\multicolumn{2}{c}{Method}                                                    & Params (M) & FLOPs (G)  & FPS & MSE $\downarrow$ & MAE $\downarrow$ & SSIM $\uparrow$ & PSNR $\uparrow$   \\ \hline
\rowcolor[HTML]{FDF0E2} 
\cellcolor[HTML]{FDF0E2}                                   & ConvLSTM       & 15.0  & 56.8  & 113 & 28.87  & 113.20  & 0.8793 & 22.07 \\
\rowcolor[HTML]{FDF0E2} 
\cellcolor[HTML]{FDF0E2}                                   & PredNet          & 12.5  & 8.4   & 659 & 185.94 & 318.30 & 0.6713 & 14.83 \\
\rowcolor[HTML]{FDF0E2} 
\cellcolor[HTML]{FDF0E2}                                   & PredRNN          & 23.8  & 116.0 & 54  & 22.01  & \textbf{91.74} & 0.9091 & 23.42 \\
\rowcolor[HTML]{FDF0E2} 
\cellcolor[HTML]{FDF0E2}                                   & PredRNN++        & 38.6  & 171.7 & 38  & \textbf{21.71}  & 91.97  & \textbf{0.9097} & \textbf{23.45} \\
\rowcolor[HTML]{FDF0E2} 
\cellcolor[HTML]{FDF0E2}                                   & MIM              & 38.0  & 179.2 & 37  & 23.09  & 96.37  & 0.9043 & 23.13 \\
\rowcolor[HTML]{FDF0E2} 
\cellcolor[HTML]{FDF0E2}                                   & E3D-LSTM         & 51.0  & 298.9 & 18  & 35.35  & 110.09  & 0.8722 & 21.27 \\
\rowcolor[HTML]{FDF0E2} 
\cellcolor[HTML]{FDF0E2}                                   & PhyDNet          & 3.1   & 15.3  & 182 & 34.75  & 125.66 & 0.8567 & 22.03 \\
\rowcolor[HTML]{FDF0E2} 
\cellcolor[HTML]{FDF0E2}                                   & MAU              & 4.5   & 17.8  & 201 & 26.56  & 104.39  & 0.8916 & 22.51 \\
\rowcolor[HTML]{FDF0E2} 
\cellcolor[HTML]{FDF0E2}                                   & PredRNNv2        & 23.9  & 116.6 & 52  & 24.13  & 97.46  & 0.9004 & 22.96 \\
\rowcolor[HTML]{FDF0E2} 
\multirow{-12}{*}{\cellcolor[HTML]{FDF0E2}Recurrent-based} & DMVFN            & 3.5   & 0.2   & 1145 & 118.32 & 220.02 & 0.7572 & 16.76 \\
\rowcolor[HTML]{E7ECE4} 
\cellcolor[HTML]{E7ECE4}                                   & SimVP            & 58.0  & 19.4  & 209 & 30.77  & 113.94  & 0.8740 & 21.81 \\
\rowcolor[HTML]{E7ECE4} 
\cellcolor[HTML]{E7ECE4}                                   & TAU              & 44.7  & 16.0  & 283 & 24.24 & 96.72 & 0.8995 & 22.87 \\
\rowcolor[HTML]{E7ECE4} 
\cellcolor[HTML]{E7ECE4}                                   & SimVPv2          & 46.8  & 16.5  & 282 & 25.86  & 101.22  & 0.8933 & 22.61 \\
\rowcolor[HTML]{E7ECE4} 
\cellcolor[HTML]{E7ECE4}                                   & ViT              & 46.1  & 16.9  & 290 & 31.05  & 115.59  & 0.8712 & 21.83 \\
\rowcolor[HTML]{E7ECE4} 
\cellcolor[HTML]{E7ECE4}                                   & Swin Transformer & 46.1  & 16.4  & 294 & 28.66  & 108.93  & 0.8815 & 22.08 \\
\rowcolor[HTML]{E7ECE4} 
\cellcolor[HTML]{E7ECE4}                                   & Uniformer        & 44.8  & 16.5  & 296 & 29.56  & 111.72  & 0.8779 & 21.97 \\
\rowcolor[HTML]{E7ECE4} 
\cellcolor[HTML]{E7ECE4}                                   & MLP-Mixer        & 38.2  & 14.7  & 334 & 28.83  & 109.51  & 0.8803 & 22.01 \\
\rowcolor[HTML]{E7ECE4} 
\cellcolor[HTML]{E7ECE4}                                   & ConvMixer        & 3.9   & 5.5   & 658 & 31.21  & 115.74  & 0.8709 & 21.71 \\
\rowcolor[HTML]{E7ECE4} 
\cellcolor[HTML]{E7ECE4}                                   & Poolformer       & 37.1  & 14.1  & 341 & 30.02  & 113.07  & 0.8750 & 21.95 \\
\rowcolor[HTML]{E7ECE4} 
\cellcolor[HTML]{E7ECE4}                                   & ConvNext         & 37.3  & 14.1  & 344 & 26.41  & 102.56  & 0.8908 & 22.49 \\
\rowcolor[HTML]{E7ECE4} 
\cellcolor[HTML]{E7ECE4}                                   & VAN              & 44.5  & 16.0  & 288 & 31.39  & 116.28  & 0.8703 & 22.82 \\
\rowcolor[HTML]{E7ECE4} 
\cellcolor[HTML]{E7ECE4}                                   & HorNet           & 45.7  & 16.3  & 287 & 29.19  & 110.17  & 0.8796 & 22.03 \\
\rowcolor[HTML]{E7ECE4} 
\multirow{-13}{*}{\cellcolor[HTML]{E7ECE4}Recurrent-free}  & MogaNet & 46.8  & 16.5  & 255 & 25.14  & 99.69  & 0.8960 & 22.73 \\
\bottomrule
\end{tabular}}\par}
\label{tab:fmmnist}
\end{table}

\begin{figure}[ht]
  \centering
  \includegraphics[width=1.0\textwidth]{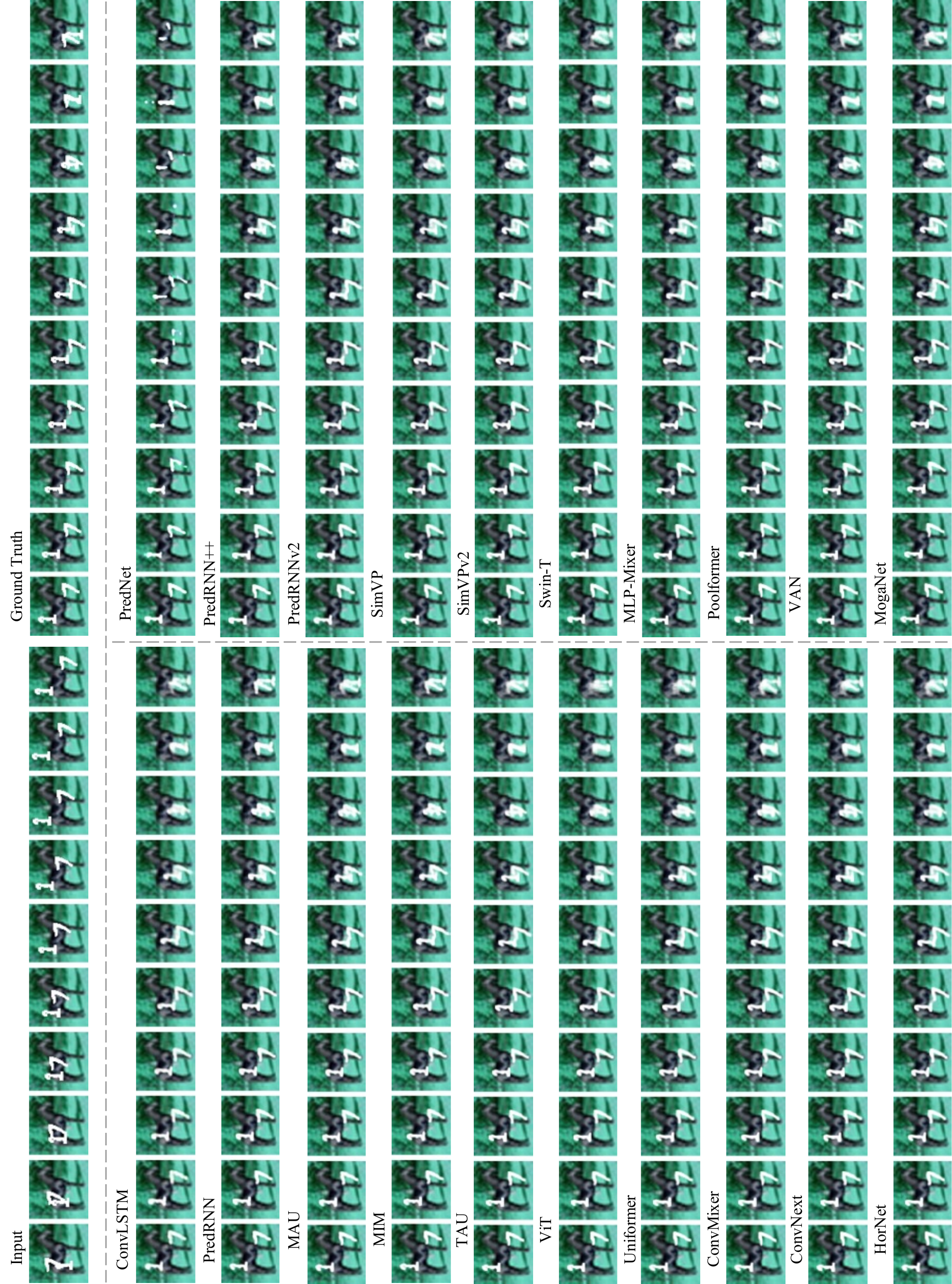}
  \caption{The qualitative visualization on Moving MNIST-CIFAR. For the convenience of formatting, we arrange the frames vertically from bottom to top.}
  \label{fig:visual_mmnist_cifar}
\end{figure}

\noindent{\textbf{Moving MNIST-CIFAR}} The quantitative results are presented in Table~\ref{tab:mmnistcifar}, while the qualitative visualizations are depicted in Figure~\ref{fig:visual_mmnist_cifar}. As the task involves more complex backgrounds, the models are required to pay greater attention to spatial modeling. Consequently, the gap between recurrent-based and recurrent-free models is narrowed.

\begin{table}[hb]
\vspace{-1mm}
\centering
\caption{The performance on the Moving MNIST-CIFAR dataset.}
{\renewcommand\baselinestretch{1.18}\selectfont
  \resizebox{\textwidth}{!}{
\begin{tabular}{ccccccccc}
\toprule
\multicolumn{2}{c}{Method}                                                    & Params (M) & FLOPs (G)  & FPS & MSE $\downarrow$ & MAE $\downarrow$ & SSIM $\uparrow$ & PSNR $\uparrow$ \\ \hline
\rowcolor[HTML]{FDF0E2} 
\cellcolor[HTML]{FDF0E2}                                   & ConvLSTM       & 15.0  & 56.8  & 113 & 73.31  & 338.56  & 0.9204 & 23.09 \\
\rowcolor[HTML]{FDF0E2} 
\cellcolor[HTML]{FDF0E2}                                   & PredNet          & 12.5  & 8.4   & 659 & 286.70 & 514.14 & 0.8139 & 17.49 \\
\rowcolor[HTML]{FDF0E2} 
\cellcolor[HTML]{FDF0E2}                                   & PredRNN          & 23.8  & 116.0 & 54  & 50.09  & 225.04 & 0.9499 & 24.90 \\
\rowcolor[HTML]{FDF0E2} 
\cellcolor[HTML]{FDF0E2}                                   & PredRNN++        & 38.6  & 171.7 & 38  & \textbf{44.19}  & 198.27  & \textbf{0.9567} & \textbf{25.60} \\
\rowcolor[HTML]{FDF0E2} 
\cellcolor[HTML]{FDF0E2}                                   & MIM              & 38.0  & 179.2 & 37  & 48.63 & 213.44 & 0.9521 & 25.08 \\
\rowcolor[HTML]{FDF0E2} 
\cellcolor[HTML]{FDF0E2}                                   & E3D-LSTM         & 51.0  & 298.9 & 18  & 80.79 & 214.86 & 0.9314 & 22.89 \\
\rowcolor[HTML]{FDF0E2} 
\cellcolor[HTML]{FDF0E2}                                   & PhyDNet          & 3.1   & 15.3  & 182 & 142.54 & 700.37 & 0.8276 & 19.92 \\
\rowcolor[HTML]{FDF0E2} 
\cellcolor[HTML]{FDF0E2}                                   & MAU              & 4.5   & 17.8  & 201 & 58.84 & 255.76 & 0.9408 & 24.19 \\
\rowcolor[HTML]{FDF0E2} 
\cellcolor[HTML]{FDF0E2}                                   & PredRNNv2        & 23.9  & 116.6 & 52  & 57.27 & 252.29 & 0.9419 & 24.24 \\
\rowcolor[HTML]{FDF0E2} 
\multirow{-12}{*}{\cellcolor[HTML]{FDF0E2}Recurrent-based} & DMVFN            & 3.5   & 0.2   & 1145 & 298.73 & 606.92 & 0.7765 & 17.07 \\
\rowcolor[HTML]{E7ECE4} 
\cellcolor[HTML]{E7ECE4}                                   & SimVP            & 58.0  & 19.4  & 209 & 59.83  & 214.54  & 0.9414 & 24.15 \\
\rowcolor[HTML]{E7ECE4} 
\cellcolor[HTML]{E7ECE4}                                   & TAU              & 44.7  & 16.0  & 283 & 48.17 & \textbf{177.35} & 0.9539 & 25.21 \\
\rowcolor[HTML]{E7ECE4} 
\cellcolor[HTML]{E7ECE4}                                   & SimVPv2          & 46.8  & 16.5  & 282 & 51.13  & 185.13 & 0.9512 & 24.93 \\
\rowcolor[HTML]{E7ECE4} 
\cellcolor[HTML]{E7ECE4}                                   & ViT              & 46.1  & 16.9  & 290 & 64.94  & 234.01 & 0.9354 & 23.90 \\
\rowcolor[HTML]{E7ECE4} 
\cellcolor[HTML]{E7ECE4}                                   & Swin Transformer & 46.1  & 16.4  & 294 & 57.11  & 207.45  & 0.9443 & 24.34 \\
\rowcolor[HTML]{E7ECE4} 
\cellcolor[HTML]{E7ECE4}                                   & Uniformer        & 44.8  & 16.5  & 296 & 56.96  & 207.51  & 0.9442 & 24.38 \\
\rowcolor[HTML]{E7ECE4} 
\cellcolor[HTML]{E7ECE4}                                   & MLP-Mixer        & 38.2  & 14.7  & 334 & 57.03  & 206.46  & 0.9446 & 24.34 \\
\rowcolor[HTML]{E7ECE4} 
\cellcolor[HTML]{E7ECE4}                                   & ConvMixer        & 3.9   & 5.5   & 658 & 59.29  & 219.76  & 0.9403 & 24.17 \\
\rowcolor[HTML]{E7ECE4} 
\cellcolor[HTML]{E7ECE4}                                   & Poolformer       & 37.1  & 14.1  & 341 & 60.98  & 219.50  & 0.9399 & 24.16 \\
\rowcolor[HTML]{E7ECE4} 
\cellcolor[HTML]{E7ECE4}                                   & ConvNext         & 37.3  & 14.1  & 344 & 51.39  & 187.17  & 0.9503 & 24.89 \\
\rowcolor[HTML]{E7ECE4} 
\cellcolor[HTML]{E7ECE4}                                   & VAN              & 44.5  & 16.0  & 288 & 59.59  & 221.32  & 0.9398 & 25.20 \\
\rowcolor[HTML]{E7ECE4} 
\cellcolor[HTML]{E7ECE4}                                   & HorNet           & 45.7  & 16.3  & 287 & 55.79  & 202.73  & 0.9456 & 24.49 \\
\rowcolor[HTML]{E7ECE4} 
\multirow{-13}{*}{\cellcolor[HTML]{E7ECE4}Recurrent-free}  & MogaNet & 46.8  & 16.5  & 255 & 49.48  & 184.11  & 0.9521 & 25.07 \\
\bottomrule
\end{tabular}}\par}
\vspace{-6mm}
\label{tab:mmnistcifar}
\end{table}

\clearpage
\clearpage

\clearpage
\clearpage

\subsection{Real-world Video Prediction}
\label{app:realworld}

\begin{figure}[ht]
  \vspace{-4mm}
  \centering
  \includegraphics[width=0.96\textwidth]{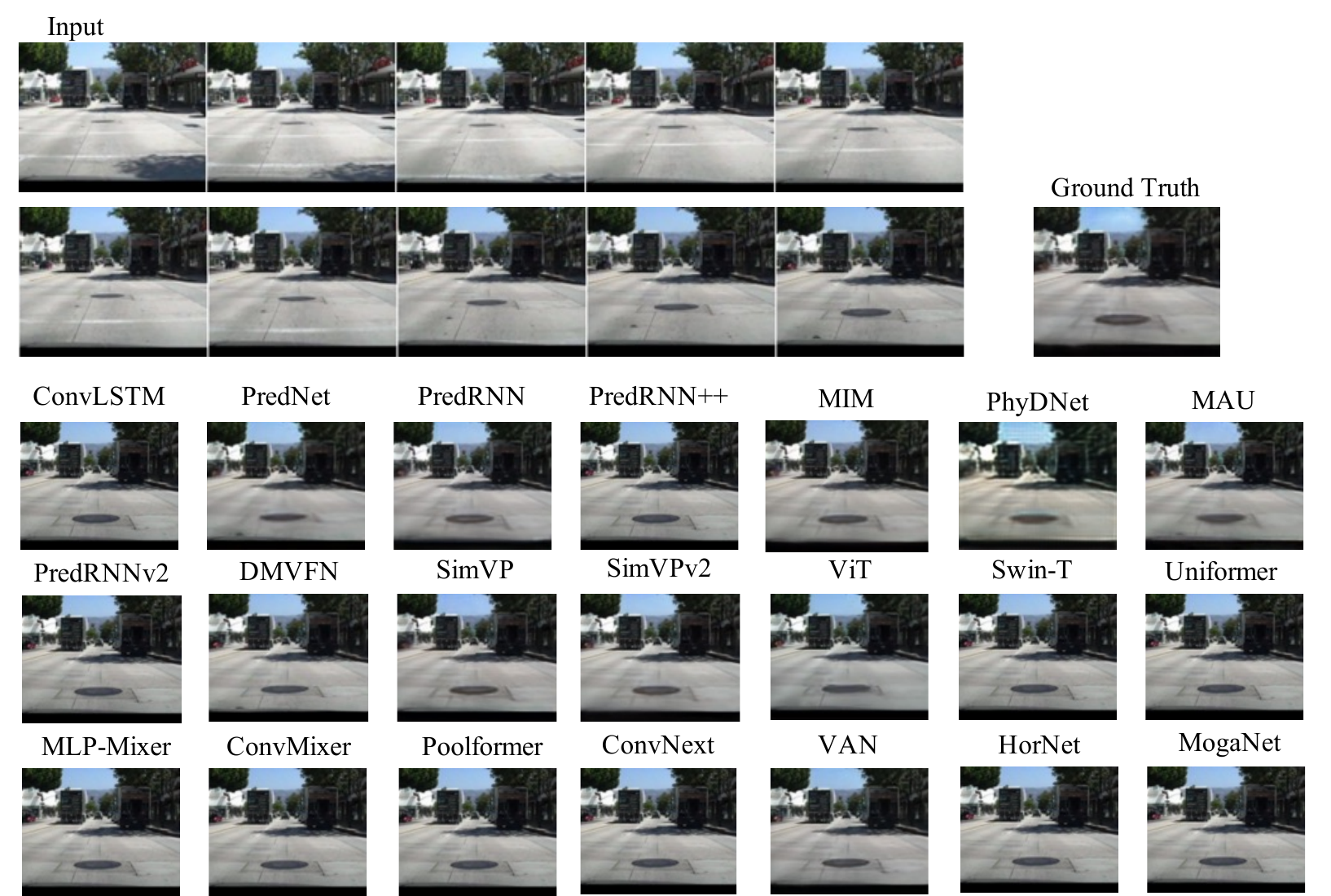}
  \vspace{-2mm}
  \caption{The qualitative visualization on Kitti\&Caltech.}
  \label{fig:visual_caltech}
  \vspace{-2mm}
\end{figure}

\noindent{\textbf{Kitties\&Caltech}} In addition to the quantitative results presented in the main text, we also provide a visualization example for qualitative assessment, as depicted in Figure~\ref{fig:visual_caltech}. Interestingly, even though PredNet and DMVFN, which have limited temporal modeling capabilities, can still perform reasonably well in predicting the next frame.

\begin{figure}[ht]
  \vspace{-6mm}
  \centering
  \includegraphics[width=0.92\textwidth]{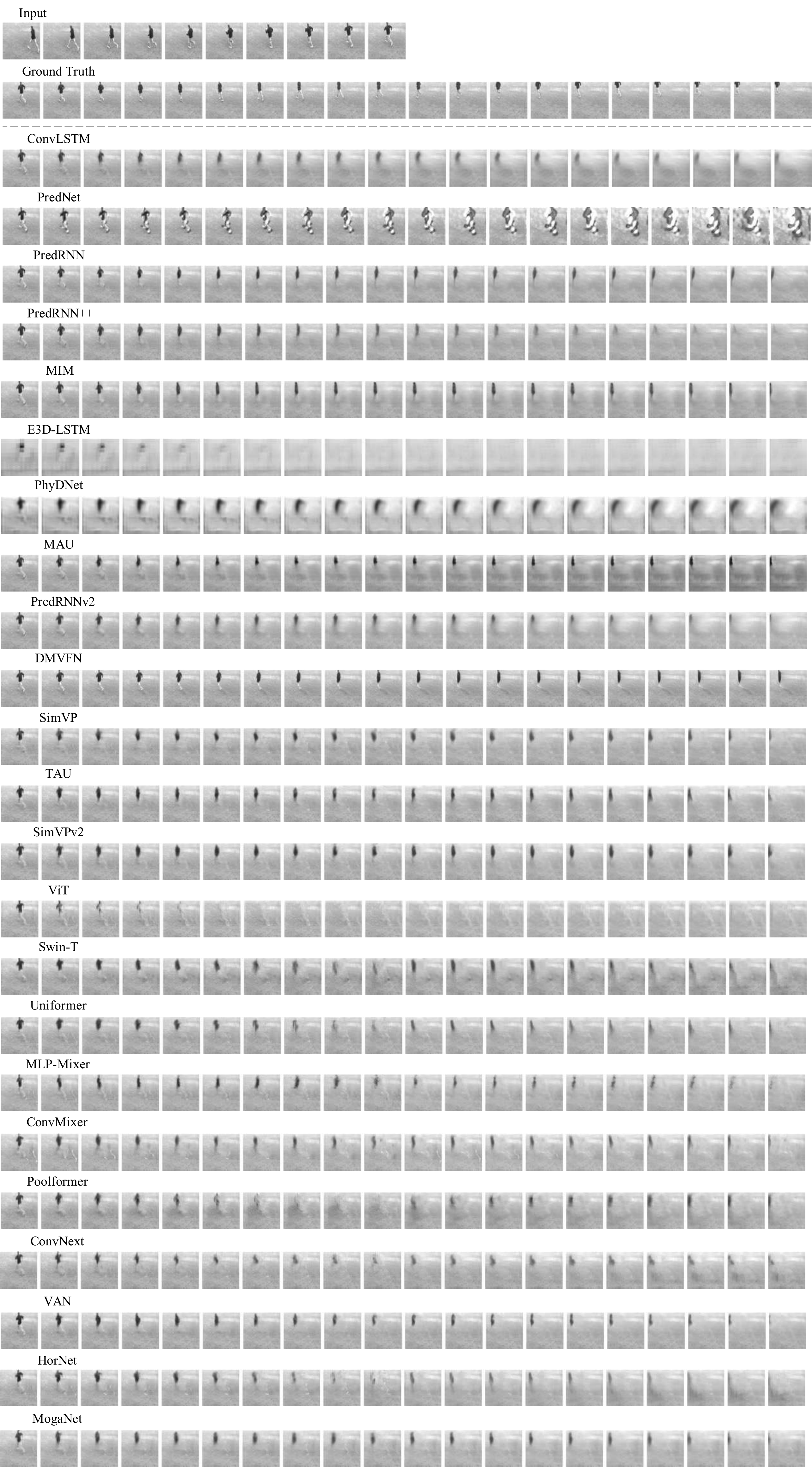}
  \caption{The qualitative visualization on KTH.}
  \label{fig:visual_kth}
\end{figure}

\noindent{\textbf{KTH}} We showcase the quantitative results and qualitative visualization on KTH in Table~\ref{tab:kth} and Figure~\ref{fig:visual_kth}, respectively. Recurrent-free models demonstrate comparable performance while requiring few computational costs, thus striking a favorable balance between performance and efficiency.

\begin{table}[ht]
  \vspace{-2mm}
  \small
  \centering
  \renewcommand\arraystretch{1.15}
  \caption{The performance on the KTH dataset.}
  \resizebox{\textwidth}{!}{%
    \begin{tabular}{cccccccccc}
      \toprule
      \multicolumn{2}{c}{Method} & Params (M) & FLOPs (G) & FPS & MSE $\downarrow$ & MAE $\downarrow$ & SSIM $\uparrow$ & PSNR $\uparrow$ & LPIPS $\downarrow$\\ \hline
      \rowcolor[HTML]{FDF0E2}
      & ConvLSTM   & 14.9   & 1368.0 & 16  & 47.65  & 445.50 & 0.8977 & 26.99 & 0.26686 \\
      \rowcolor[HTML]{FDF0E2}
      & PredNet    & 12.5   & 3.4    & 399 & 152.11 & 783.10 & 0.8094 & 22.45 & 0.32159 \\
      \rowcolor[HTML]{FDF0E2}
      & PredRNN    & 23.6   & 2800.0 & 7   & 41.07  & 380.60 & 0.9097 & 27.95 & 0.21892 \\
      \rowcolor[HTML]{FDF0E2}
      & PredRNN++  & 38.3   & 4162.0 & 5   & \textbf{39.84} & \textbf{370.40} & \textbf{0.9124} & \textbf{28.13} & 0.19871 \\
      \rowcolor[HTML]{FDF0E2}
      & MIM        & 39.8   & 1099.0 & 17  & 40.73  & 380.80 & 0.9025 & 27.78 & 0.18808 \\
      \rowcolor[HTML]{FDF0E2}
      & E3D-LSTM   & 53.5   & 217.0  & 17  & 136.40 & 892.70 & 0.8153 & 21.78 & 0.48358 \\
      \rowcolor[HTML]{FDF0E2}
      & PhyDNet    & 3.1    & 93.6   & 58  & 91.12  & 765.60 & 0.8322 & 23.41 & 0.50155 \\
      \rowcolor[HTML]{FDF0E2}
      & MAU        & 20.1   & 399.0  & 8   & 51.02  & 471.20 & 0.8945 & 26.73 & 0.25442 \\
      \rowcolor[HTML]{FDF0E2}
      & PredRNNv2  & 23.6   & 2815.0  & 7   & 39.57  & 368.80 & 0.9099 & 28.01 & 0.21478 \\
      \rowcolor[HTML]{FDF0E2}
      \rowcolor[HTML]{FDF0E2}
      \multirow{-7}{*}{\cellcolor[HTML]{FDF0E2}Recurrent-based} & DMVFN      & 3.5    & 0.9    & 727 & 59.61  & 413.20 & 0.8976 & 26.65 & \textbf{0.12842} \\
      \rowcolor[HTML]{E7ECE4}
      & SimVP      & 12.2   & 62.8   & 77  & 41.11  & 397.10 & 0.9065 & 27.46 & 0.26496 \\
      \rowcolor[HTML]{E7ECE4}
      & TAU        & 15.0   & 73.8   & 55  & 45.32  & 421.70 & 0.9086 & 27.10 & 0.22856 \\
      \rowcolor[HTML]{E7ECE4}
      & SimVPv2    & 15.6   & 76.8   & 53  & 45.02  & 417.80 & 0.9049 & 27.04 & 0.25240 \\
      \rowcolor[HTML]{E7ECE4}
      & ViT        & 12.7   & 112.0  & 28  & 56.57  & 459.30 & 0.8947 & 26.19 & 0.27494 \\
      \rowcolor[HTML]{E7ECE4}
      & Swin-T     & 15.3   & 75.9   & 65  & 45.72  & 405.70 & 0.9039 & 27.01 & 0.25178 \\
      \rowcolor[HTML]{E7ECE4}
      & Uniformer  & 11.8   & 78.3   & 43  & 44.71  & 404.60 & 0.9058 & 27.16 & 0.24174 \\
      \rowcolor[HTML]{E7ECE4}
      & MLP-Mixer  & 20.3   & 66.6   & 34  & 57.74  & 517.40 & 0.8886 & 25.72 & 0.28799 \\
      \rowcolor[HTML]{E7ECE4}
      & ConvMixer  & 1.5    & 18.3   & 175 & 47.31  & 446.10 & 0.8993 & 26.66 & 0.28149 \\
      \rowcolor[HTML]{E7ECE4}
      & Poolformer & 12.4   & 63.6   & 67  & 45.44  & 400.90 & 0.9065 & 27.22 & 0.24763 \\
      \rowcolor[HTML]{E7ECE4}
      & ConvNext   & 12.5   & 63.9   & 72  & 45.48  & 428.30 & 0.9037 & 26.96 & 0.26253 \\
      \rowcolor[HTML]{E7ECE4}
      & VAN        & 14.9   & 73.8   & 55  & 45.05  & 409.10 & 0.9074 & 27.07 & 0.23116 \\
      \rowcolor[HTML]{E7ECE4}
      & HorNet     & 15.3   & 75.3   & 58  & 46.84  & 421.20 & 0.9005 & 26.80 & 0.26921 \\
      \rowcolor[HTML]{E7ECE4}
      \multirow{-13}{*}{\cellcolor[HTML]{E7ECE4}Recurrent-free} & MogaNet    & 15.6   & 76.7   & 48  & 42.98  & 418.70  & 0.9065 & 27.16 & 0.25146 \\
      \bottomrule
    \end{tabular}%
  }
  \label{tab:kth}
  \vspace{-2mm}
\end{table}

\clearpage
\clearpage

\begin{figure}[ht]
  \vspace{8mm}
  \centering
  \includegraphics[width=0.88\textwidth]{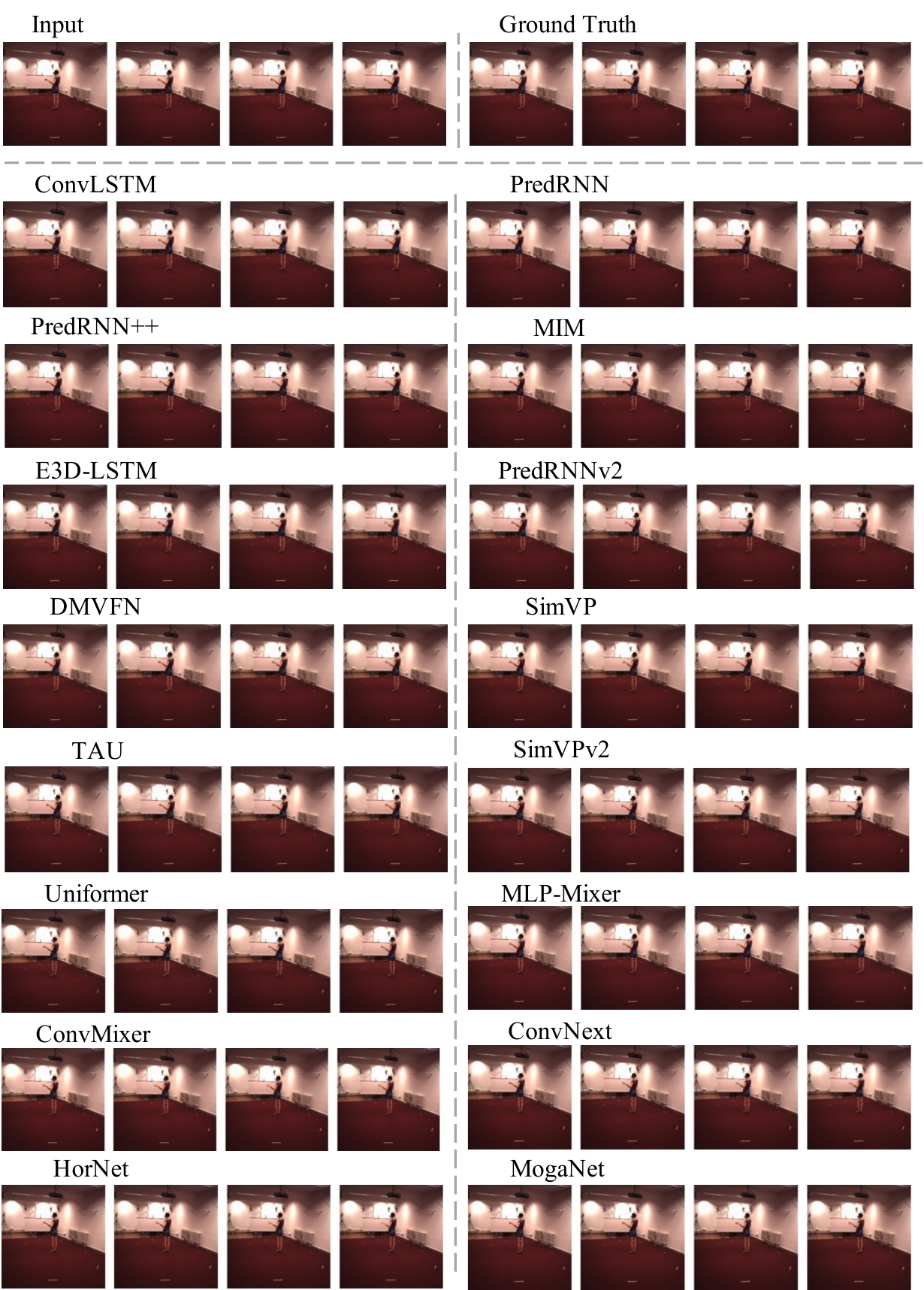}
  \caption{The qualitative visualization on Human3.6M.}
  \label{fig:visual_human}
  % \vspace{-4mm}
\end{figure}

\noindent{\textbf{Human3.6M}} The quantitative results are presented in Table~\ref{tab:human}, and the qualitative visualization is depicted in Figure~\ref{fig:visual_human}. In this task, human motion exhibits subtle changes between adjacent frames, resulting in a low-frequency signal of overall dynamics. Consequently, recurrent-free models, which excel at spatial learning, can efficiently and accurately predict future frames.

\begin{table}[ht]
  \vspace{-2mm}
  \small
  \centering
  \renewcommand\arraystretch{1.1}
  \caption{The performance on the Human3.6M dataset.}
  \resizebox{\textwidth}{!}{%
    \begin{tabular}{cccccccccc}
      \toprule
      \multicolumn{2}{c}{Method} & Params (M) & FLOPs (G) & FPS & MSE $\downarrow$ & MAE $\downarrow$ & SSIM $\uparrow$ & PSNR $\uparrow$ & LPIPS $\downarrow$\\ \hline
      \rowcolor[HTML]{FDF0E2}
      & ConvLSTM   & 15.5   & 347.0 & 52  & 125.5  & 1566.7 & 0.9813 & 33.40 & 0.03557 \\
      \rowcolor[HTML]{FDF0E2}
      & PredNet   & 12.5   & 13.7  & 176 & 261.9  & 1625.3 & 0.9786 & 31.76 & 0.03264 \\
      \rowcolor[HTML]{FDF0E2}
      & PredRNN    & 24.6  & 704.0 & 25  & 113.2 & 1458.3 & 0.9831 & 33.94 & 0.03245 \\
      \rowcolor[HTML]{FDF0E2}
      & PredRNN++  & 39.3  & 1033.0 & 18 & 110.0 & 1452.2 & 0.9832 & 34.02 & 0.03196 \\
      \rowcolor[HTML]{FDF0E2}
      & MIM        & 47.6  & 1051.0 & 17 & 112.1 & 1467.1 & 0.9829 & 33.97 & 0.03338 \\
      \rowcolor[HTML]{FDF0E2}
      & E3D-LSTM   & 60.9   & 542.0  & 7 & 143.3 & 1442.5 & 0.9803 & 32.52 & 0.04133 \\
      \rowcolor[HTML]{FDF0E2}
      & PhyDNet    & 4.2    & 19.1   & 57 & 125.7 & 1614.7 & 0.9804 & 33.05 & 0.03709 \\
      \rowcolor[HTML]{FDF0E2}
      & MAU        & 20.2   & 105.0  & 6  & 127.3 & 1577.0 & 0.9812 & 33.33 & 0.03561 \\
      \rowcolor[HTML]{FDF0E2}
      & PredRNNv2  & 24.6   & 708.0  & 24 & 114.9 & 1484.7 & 0.9827 & 33.84 & 0.03334 \\
      \rowcolor[HTML]{FDF0E2}
      \rowcolor[HTML]{FDF0E2}
      \multirow{-7}{*}{\cellcolor[HTML]{FDF0E2}Recurrent-based} & DMVFN      & 8.6    & 63.6 & 341 & 109.3  & 1449.3 & 0.9833 & 34.05 & 0.03189 \\
      \rowcolor[HTML]{E7ECE4}
      & SimVP      & 41.2   & 197.0 & 26 & 115.8  & 1511.5 & 0.9822 & 33.73 & 0.03467 \\
      \rowcolor[HTML]{E7ECE4}
      & TAU        & 37.6 & 182.0 & 26 & 113.3 & \textbf{1390.7} & \textbf{0.9839} & 34.03 & \textbf{0.02783} \\
      \rowcolor[HTML]{E7ECE4}
      & SimVPv2    & 11.3   & 74.6 & 52 & 108.4 & 1441.0 & 0.9834 & \textbf{34.08} & 0.03224 \\
      \rowcolor[HTML]{E7ECE4}
      & ViT        & 28.3   & 239.0 & 17 & 136.3 & 1603.5 & 0.9796 & 33.10 & 0.03729 \\
      \rowcolor[HTML]{E7ECE4}
      & Swin       & 38.8   & 188.0 & 28 & 133.2 & 1599.7 & 0.9799 & 33.16 & 0.03766 \\
      \rowcolor[HTML]{E7ECE4}
      & Uniformer  & 27.7  & 211.0 & 14 & 116.3 & 1497.7 & 0.9824 & 33.76 & 0.03385 \\
      \rowcolor[HTML]{E7ECE4}
      & MLP-Mixer  & 47.0  & 164.0 & 34 & 125.7 & 1511.9 & 0.9819 & 33.49 & 0.03417 \\
      \rowcolor[HTML]{E7ECE4}
      & ConvMixer  & 3.1   & 39.4  & 84 & 115.8  & 1527.4 & 0.9822 & 33.67 & 0.03436 \\
      \rowcolor[HTML]{E7ECE4}
      & Poolformer & 31.2  & 156.0 & 30 & 118.4  & 1484.1 & 0.9827 & 33.78 & 0.03313 \\
      \rowcolor[HTML]{E7ECE4}
      & ConvNext   & 31.4   & 157.0 & 33 & 113.4 & 1469.7 & 0.9828 & 33.86 & 0.03305 \\
      \rowcolor[HTML]{E7ECE4}
      & VAN        & 37.5   & 182.0 & 24 & 111.4 & 1454.5 & 0.9831 & 33.93 & 0.03335 \\
      \rowcolor[HTML]{E7ECE4}
      & HorNet     & 28.1  & 143.0 & 33 & 118.1 & 1481.1 & 0.9825 & 33.73 & 0.03333 \\
      \rowcolor[HTML]{E7ECE4}
      \multirow{-13}{*}{\cellcolor[HTML]{E7ECE4}Recurrent-free} & MogaNet   & 8.6  & 163.6 & 56	& 109.1 & 1446.4 & 0.9834 & 34.05 & 0.03163 \\
      \bottomrule
    \end{tabular}%
  }
  \label{tab:human}
  \vspace{-2mm}
\end{table}

\subsection{Traffic and Weather Forecasting}
\label{app:trafficweather} 

\subsubsection{TaxiBJ}

\begin{figure}[ht]
  \vspace{-4mm}
  \centering
  \includegraphics[width=0.88\textwidth]{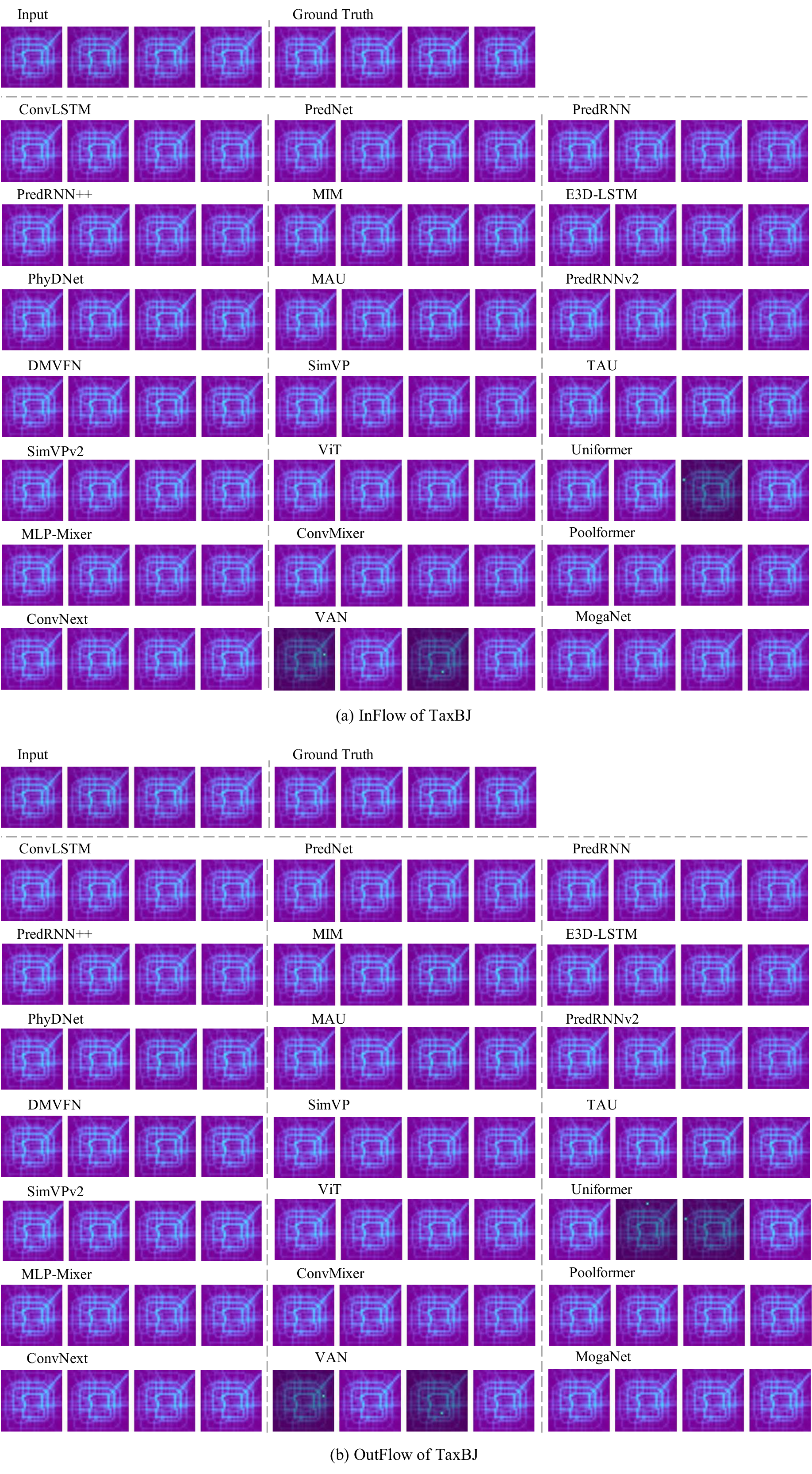}
  \caption{The qualitative visualization on TaxiBJ.}
  \label{fig:visual_taxibj}
  \vspace{-4mm}
\end{figure}

We show the quantitative results in Table~\ref{tab:taxibj} and qualitative visualizations in Figure~\ref{fig:visual_taxibj}. The recurrent-free models have shown promising results in low-frequency traffic flow data than their counterparts.

\begin{table}[ht]
  \vspace{-2mm}
  \small
  \centering
  \renewcommand\arraystretch{1.00}
  \caption{The performance on the TaxiBJ dataset.}
  \resizebox{\textwidth}{!}{%
    \begin{tabular}{cccccccc}
      \toprule
      \multicolumn{2}{c}{Method} & Params (M) & FLOPs (G) & FPS & MSE $\downarrow$ & MAE $\downarrow$ & SSIM $\uparrow$\\ \hline
      \rowcolor[HTML]{FDF0E2}
      & ConvLSTM & 15.0 & 20.7 & 815 & 0.3358 & 15.32 & 0.9836 \\
      \rowcolor[HTML]{FDF0E2}
      & PredNet & 12.5 & 0.9 & 5031 & 0.3516 & 15.91 & 0.9828 \\
      \rowcolor[HTML]{FDF0E2}
      & PredRNN & 23.7 & 42.4 & 416 & 0.3194 & 15.31 & 0.9838 \\
      \rowcolor[HTML]{FDF0E2}
      & PredRNN++ & 38.4 & 63.0 & 301 & 0.3348 & 15.37 & 0.9834  \\
      \rowcolor[HTML]{FDF0E2}
      & MIM & 37.9 & 64.1 & 275 & 0.3110 & 14.96 & 0.9847 \\
      \rowcolor[HTML]{FDF0E2}
      & E3DLSTM & 51.0 & 98.19 & 60 & 0.3421 & 14.98 & 0.9842 \\
      \rowcolor[HTML]{FDF0E2}
      & PhyDNet & 3.1 & 5.6 & 982 & 0.3622 & 15.53 & 0.9828  \\
      \rowcolor[HTML]{FDF0E2}
      & MAU & 4.4 & 6.0 & 540 & 0.3268 & 15.26 & 0.9834 \\
      \rowcolor[HTML]{FDF0E2}
      & PredRNNv2 & 23.7 & 42.6 & 378 & 0.3834 & 15.55 & 0.9826  \\
      \rowcolor[HTML]{FDF0E2}
      \multirow{-12}{*}{\cellcolor[HTML]{FDF0E2}Recurrent-based} & DMVFN & 3.5 & 57.1 & 4772 & 0.3517 & 15.72 & 0.9833  \\
      \rowcolor[HTML]{E7ECE4}
      & SimVP & 13.8 & 3.6 & 533 & 0.3282 & 15.45 & 0.9835 \\
      \rowcolor[HTML]{E7ECE4}
       & TAU & 9.6 & 2.5 & 1268 & 0.3108 & 14.93 & \textbf{0.9848} \\
      \rowcolor[HTML]{E7ECE4}
      & SimVPv2 & 10.0 & 2.6 & 1217 & 0.3246 & 15.03 & 0.9844 \\
      \rowcolor[HTML]{E7ECE4}
      & ViT & 9.7 & 2.8 & 1301 & 0.3171 & 15.15 & 0.9841 \\
      \rowcolor[HTML]{E7ECE4}
      & Swin Transformer & 9.7 & 2.6 & 1506 & 0.3128 & 15.07 & 0.9847 \\
      \rowcolor[HTML]{E7ECE4}
      & Uniformer & 9.5 & 2.7 & 1333 & 0.3268 & 15.16 & 0.9844 \\
      \rowcolor[HTML]{E7ECE4}
      & MLP-Mixer & 8.2 & 2.2 & 1974 & 0.3206 & 15.37 & 0.9841 \\
      \rowcolor[HTML]{E7ECE4}
      & ConvMixer & 0.8 & 0.2 & 4793 & 0.3634 & 15.63 & 0.9831 \\
      \rowcolor[HTML]{E7ECE4}
      & Poolformer & 7.6 & 2.1 & 1827 & 0.3273 & 15.39 & 0.9840\\
      \rowcolor[HTML]{E7ECE4}
      & ConvNext & 7.8 & 2.1 & 1918 & \textbf{0.3106} & \textbf{14.90} & 0.9845 \\
      \rowcolor[HTML]{E7ECE4}
      & VAN & 9.5 & 2.5 & 1273 & 0.3125 & 14.96 & \textbf{0.9848} \\
      \rowcolor[HTML]{E7ECE4}
      & HorNet & 9.7 & 2.5 & 1350 & 0.3186 & 15.01 & 0.9843 \\
      \rowcolor[HTML]{E7ECE4}
      \multirow{-13}{*}{\cellcolor[HTML]{E7ECE4}Recurrent-free} & MogaNet & 10.0 & 2.6 & 1005 & 0.3114 & 15.06 & 0.9847 \\
      \bottomrule
    \end{tabular}%
  }
  \label{tab:taxibj}
  \vspace{-2mm}
\end{table}

\clearpage
\clearpage

\clearpage
\clearpage

\subsubsection{WeatherBench}
\label{app:weatherbench}

\textit{We strongly recommend readers refer to the GIF animations provided in our GitHub (\href{https://github.com/chengtan9907/OpenSTL/tree/master/docs/en/visualization}{OpenSTL/docs/en/visualization/}), as they provide a clearer visualization of the model's prediction performance.}

\begin{figure}[ht]
  % \vspace{-4mm}
  \centering
  \includegraphics[width=1.0\textwidth]{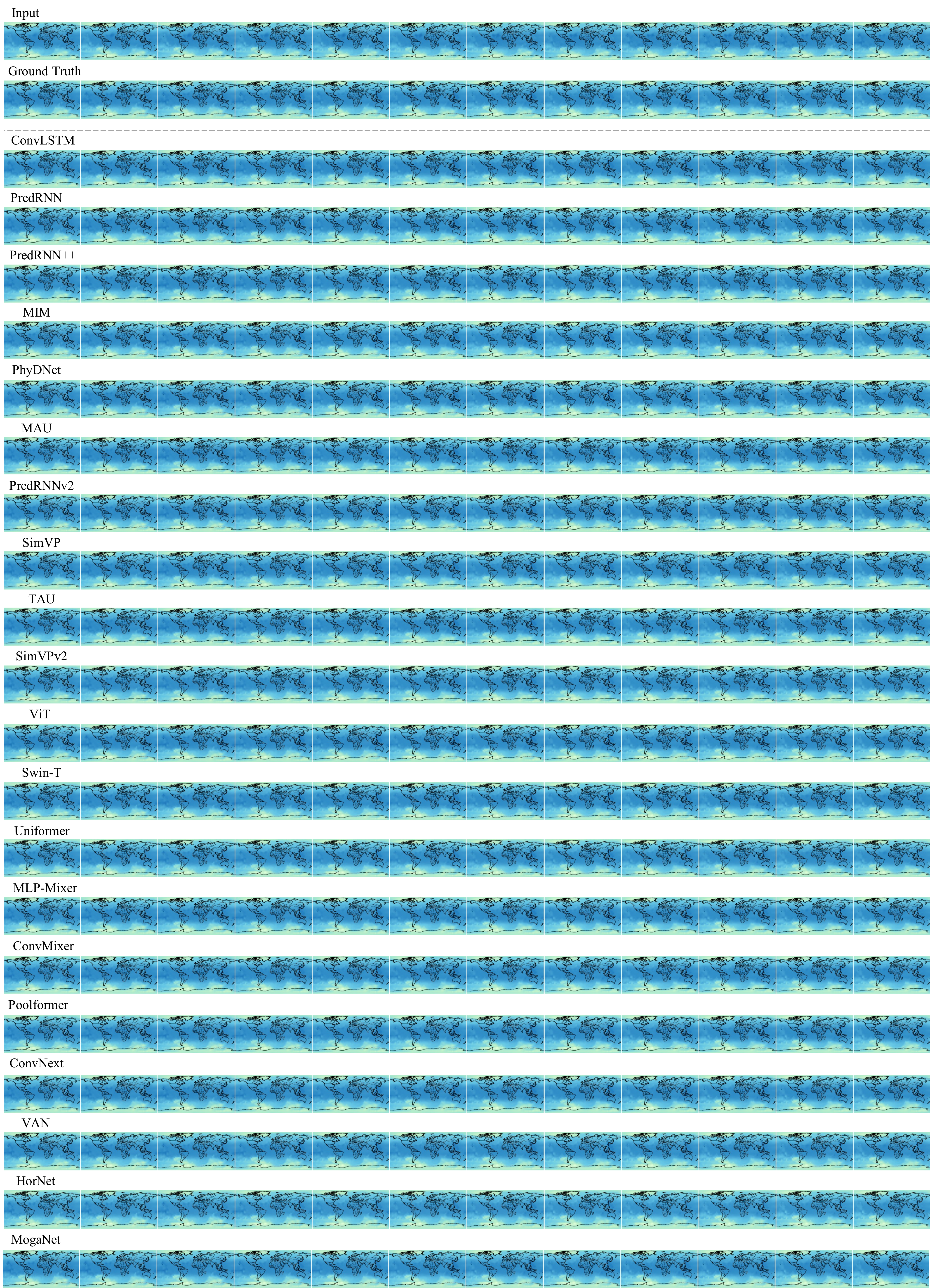}
  \caption{The qualitative visualization on the single-variable temperature forecasting in the WeatherBench dataset.}
  \label{fig:visual_weather_t2m}
  % \vspace{-4mm}
\end{figure}

\noindent{\textbf{Single-variable Temperature Forecasting}} The quantitative results and qualitative visualization are presented in Table~\ref{tab:single_temp} and Figure~\ref{fig:visual_weather_t2m}. The recurrent-free models exhibit a clear superiority over the recurrent-based models in terms of both performance and efficiency, achieving a landslide victory.

\begin{table}[ht]
\vspace{-1mm}
\small
\centering
\renewcommand\arraystretch{1.15}
\caption{The performance on the single-variable temperature forecasting in WeatherBench.}
\resizebox{\textwidth}{!}{%
  \begin{tabular}{ccccccccc}
    \toprule
    \multicolumn{2}{c}{Method} & Params (M) & FLOPs (G) & FPS & MSE $\downarrow$ & MAE $\downarrow$ & RMSE $\downarrow$  \\ \hline
    \rowcolor[HTML]{FDF0E2}
    & ConvLSTM & 14.9 & 136.0 & 46 & 1.521 & 0.7949 & 1.233 \\
    \rowcolor[HTML]{FDF0E2}
    & PredRNN & 23.6 & 278.0 & 22 & 1.331 & 0.7246 & 1.154 \\
    \rowcolor[HTML]{FDF0E2}
    & PredRNN++ & 38.3 & 413.0 & 15 & 1.634 & 0.7883 & 1.278 \\
    \rowcolor[HTML]{FDF0E2}
    & MIM & 37.8 & 109.0 & 126 & 1.784 & 0.8716 & 1.336 \\
    \rowcolor[HTML]{FDF0E2}
    & PhyDNet & 3.1 & 36.8 & 177 & 285.9 & 8.7370 & 16.91 \\
    \rowcolor[HTML]{FDF0E2}
    & MAU & 5.5 & 39.6 & 237 & 1.251 & 0.7036 & 1.119 \\
    \rowcolor[HTML]{FDF0E2}
    \multirow{-7}{*}{\cellcolor[HTML]{FDF0E2}Recurrent-based} & PredRNNv2 & 23.6 & 279.0 & 22 & 1.545 & 0.7986 & 1.243 \\
    \rowcolor[HTML]{E7ECE4}
    & SimVP & 14.7 & 8.0 & 160 & 1.238 & 0.7037 & 1.113 \\
    \rowcolor[HTML]{E7ECE4}
    & TAU & 12.2 & 6.7 & 511 & 1.162 & 0.6707 & 1.078 \\
    \rowcolor[HTML]{E7ECE4}
    & SimVPv2 & 12.8 & 7.0 & 504 & \textbf{1.105} & \textbf{0.6567} & \textbf{1.051} \\
    \rowcolor[HTML]{E7ECE4}
    & ViT & 12.4 & 8.0 & 432 & 1.146 & 0.6712 & 1.070 \\
    \rowcolor[HTML]{E7ECE4}
    & Swin Transformer & 12.4 & 6.9 & 581 & 1.143 & 0.6735, & 1.069 \\
    \rowcolor[HTML]{E7ECE4}
    & Uniformer & 12.0 & 7.5 & 465 & 1.204 & 0.6885 & 1.097 \\
    \rowcolor[HTML]{E7ECE4}
    & MLP-Mixer & 11.1 & 5.9 & 713 & 1.255 & 0.7011 & 1.119 \\
    \rowcolor[HTML]{E7ECE4}
    & ConvMixer & 1.1 & 1.0 & 1705 & 1.267 & 0.7073 & 1.126 \\
    \rowcolor[HTML]{E7ECE4}
    & Poolformer & 10.0 & 5.6 & 722 & 1.156 & 0.6715 & 1.075 \\
    \rowcolor[HTML]{E7ECE4}
    & ConvNext & 10.1 & 5.7 & 689 & 1.277 & 0.7220 & 1.130 \\
    \rowcolor[HTML]{E7ECE4}
    & VAN & 12.2 & 6.7 & 523 & 1.150 & 0.6803 & 1.072 \\
    \rowcolor[HTML]{E7ECE4}
    & HorNet & 12.4 & 6.8 & 517 & 1.201 & 0.6906 & 1.096 \\
    \rowcolor[HTML]{E7ECE4}
    \multirow{-13}{*}{\cellcolor[HTML]{E7ECE4}Recurrent-free} & MogaNet & 12.8 & 7.0 & 416 & 1.152 & 0.6665 & 1.073 \\
    \bottomrule
  \end{tabular}%
}
\vspace{-1mm}
\label{tab:single_temp}
\end{table}

\begin{figure}[ht]
  % \vspace{-4mm}
  \centering
  \includegraphics[width=1.0\textwidth]{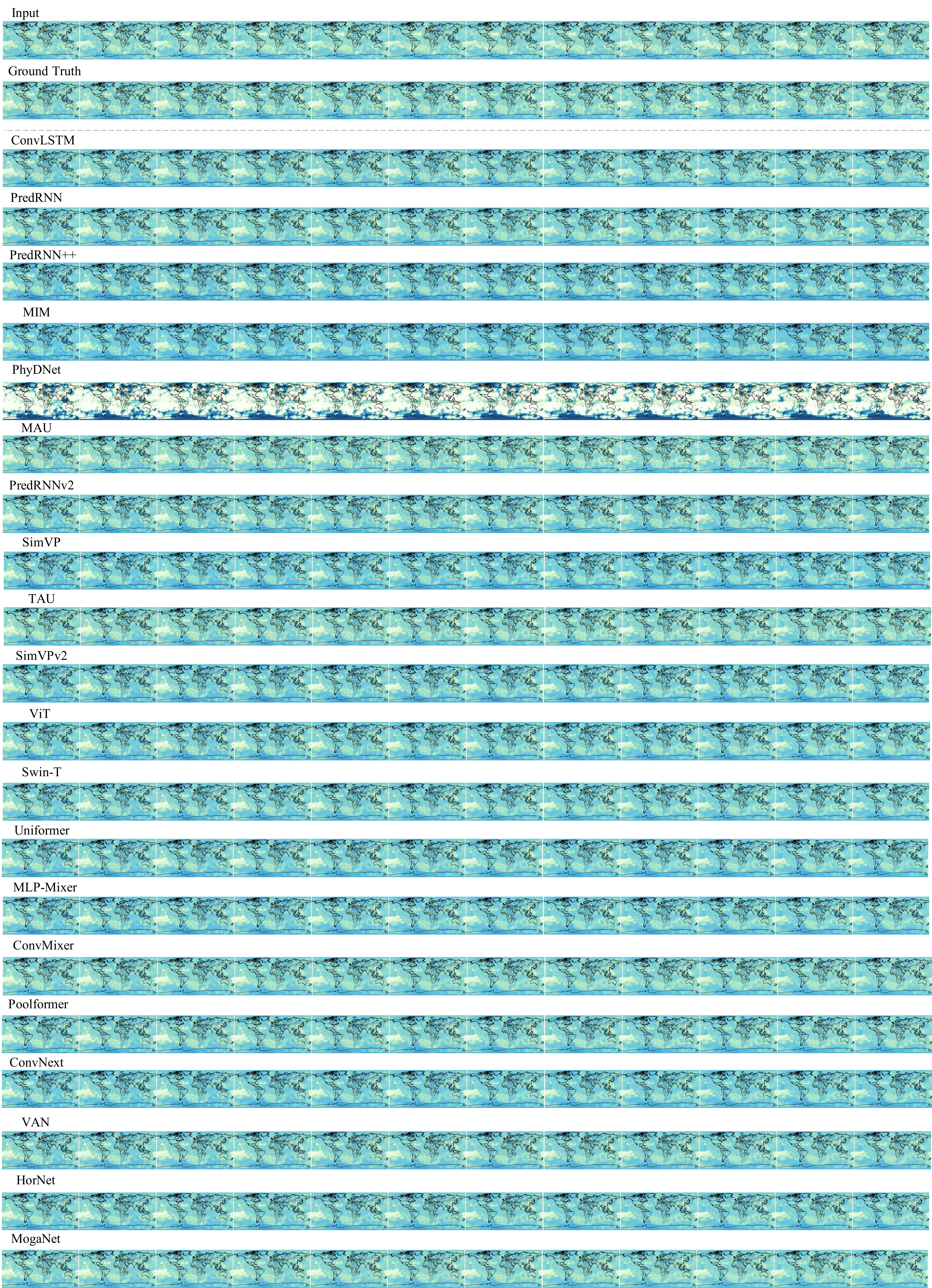}
  \caption{The qualitative visualization on the single-variable humidity forecasting in the WeatherBench dataset.}
  \label{fig:visual_weather_r}
  % \vspace{-4mm}
\end{figure}

\noindent{\textbf{Single-variable Humidity Forecasting}} The quantitative results and qualitative visualization are presented in Table~\ref{tab:single_humidity} and Figure~\ref{fig:visual_weather_r}. The results are almost consistent with the temperature forecasting.

\begin{table}[hb]
\vspace{-1mm}
\small
\centering
\renewcommand\arraystretch{1.15}
\caption{The performance on the single-variable humidity forecasting in WeatherBench.}
\resizebox{\textwidth}{!}{%
  \begin{tabular}{ccccccccc}
    \toprule
    \multicolumn{2}{c}{Method} & Params (M) & FLOPs (G) & FPS & MSE $\downarrow$ & MAE $\downarrow$ & RMSE $\downarrow$  \\ \hline
    \rowcolor[HTML]{FDF0E2}
    & ConvLSTM & 14.9 & 136.0 & 46 & 35.146 & 4.012 & 5.928 \\
    \rowcolor[HTML]{FDF0E2}
    & PredRNN & 23.6 & 278.0 & 22 & 37.611 & 4.096 & 6.133 \\
    \rowcolor[HTML]{FDF0E2}
    & PredRNN++ & 38.3 & 413.0 & 15 & 45.993 & 4.731 & 6.782 \\
    \rowcolor[HTML]{FDF0E2}
    & MIM & 37.8 & 109.0 & 126 & 61.113 & 5.504 & 7.817 \\
    \rowcolor[HTML]{FDF0E2}
    & PhyDNet & 3.1 & 36.8 & 177 & 239.0 & 8.975 & 15.46 \\
    \rowcolor[HTML]{FDF0E2}
    & MAU & 5.5 & 39.6 & 237 & 34.529 & 4.004 & 5.876 \\
    \rowcolor[HTML]{FDF0E2}
    \multirow{-7}{*}{\cellcolor[HTML]{FDF0E2}Recurrent-based} & PredRNNv2 & 23.6 & 279.0 & 22 & 36.508 & 4.087 & 6.042 \\
    \rowcolor[HTML]{E7ECE4}
    & SimVP & 14.7 & 8.0 & 160 & 34.355 & 3.994 & 5.861 \\
    \rowcolor[HTML]{E7ECE4}
    & TAU & 12.2 & 6.7 & 511 & 31.831 & 3.818 & 5.642 \\
    \rowcolor[HTML]{E7ECE4}
    & SimVPv2 & 12.8 & 7.0 & 504 & 31.426 & \textbf{3.765} & 5.606 \\
    \rowcolor[HTML]{E7ECE4}
    & ViT & 12.4 & 8.0 & 432 & 32.616 & 3.852 & 5.711 \\
    \rowcolor[HTML]{E7ECE4}
    & Swin Transformer & 12.4 & 6.9 & 581 & \textbf{31.332} & 3.776 & \textbf{5.597} \\
    \rowcolor[HTML]{E7ECE4}
    & Uniformer & 12.0 & 7.5 & 465 & 32.199 & 3.864 & 5.674 \\
    \rowcolor[HTML]{E7ECE4}
    & MLP-Mixer & 11.1 & 5.9 & 713 & 34.467 & 3.950 & 5.871 \\
    \rowcolor[HTML]{E7ECE4}
    & ConvMixer & 1.1 & 1.0 & 1705 & 32.829 & 3.909 & 5.730 \\
    \rowcolor[HTML]{E7ECE4}
    & Poolformer & 10.0 & 5.6 & 722 & 31.989 & 3.803 & 5.656 \\
    \rowcolor[HTML]{E7ECE4}
    & ConvNext & 10.1 & 5.7 & 689 & 33.179 & 3.928 & 5.760 \\
    \rowcolor[HTML]{E7ECE4}
    & VAN & 12.2 & 6.7 & 523 & 31.712 & 3.812 & 5.631 \\
    \rowcolor[HTML]{E7ECE4}
    & HorNet & 12.4 & 6.8 & 517 & 32.081 & 3.826 & 5.664 \\
    \rowcolor[HTML]{E7ECE4}
    \multirow{-13}{*}{\cellcolor[HTML]{E7ECE4}Recurrent-free} & MogaNet & 12.8 & 7.0 & 416 & 31.795 & 3.816 & 5.639 \\
    \bottomrule
  \end{tabular}%
}
\label{tab:single_humidity}
\vspace{-2mm}
\end{table}

\clearpage

\clearpage

\begin{figure}[ht]
  % \vspace{-4mm}
  \centering
  \includegraphics[width=1.0\textwidth]{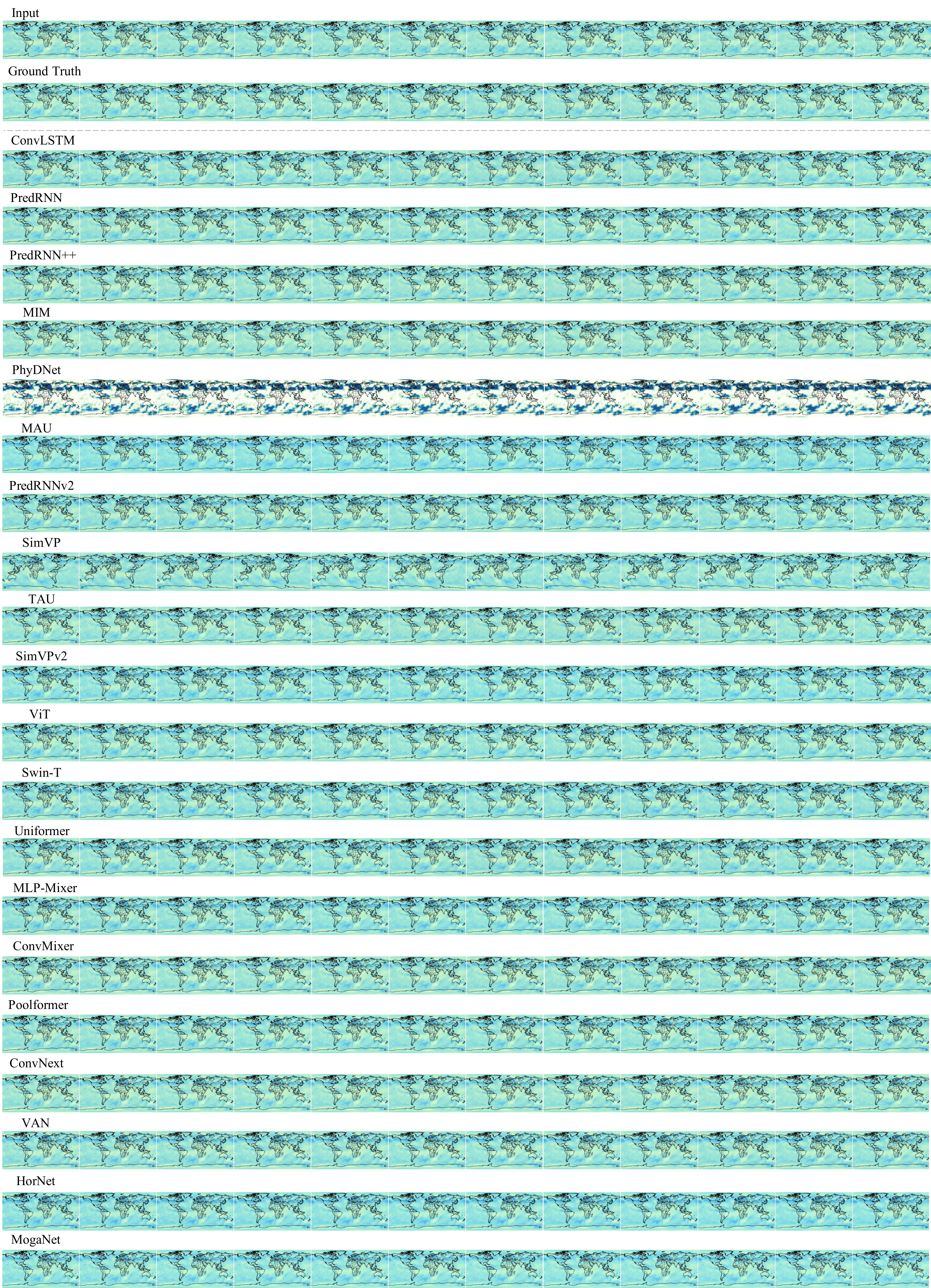}
  \caption{The qualitative visualization on the single-variable latitude wind forecasting in the WeatherBench dataset.}
  \label{fig:visual_weather_u}
  % \vspace{-4mm}
\end{figure}

\begin{figure}[ht]
  % \vspace{-4mm}
  \centering
  \includegraphics[width=1.0\textwidth]{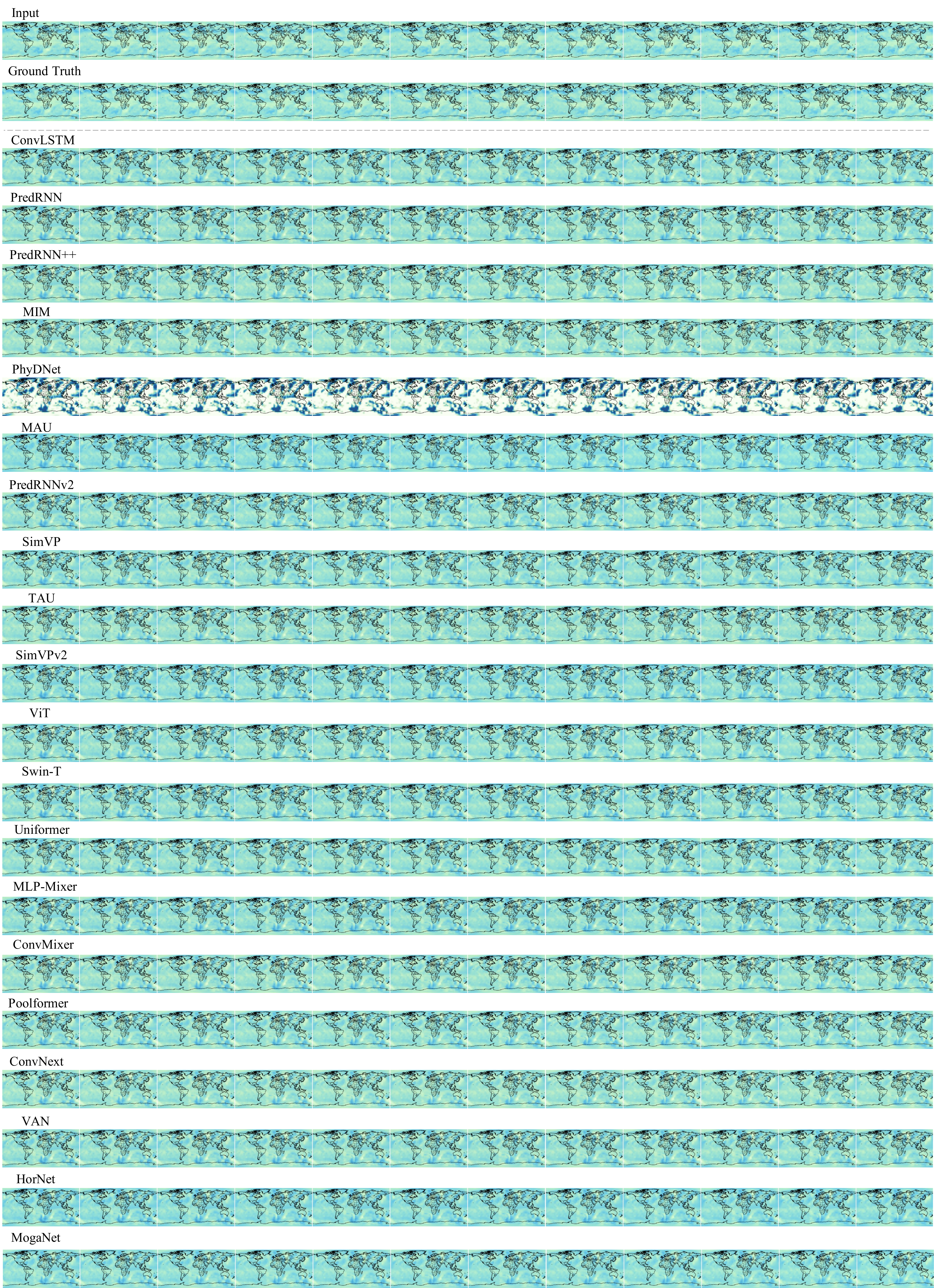}
  \caption{The qualitative visualization on the single-variable longitude wind forecasting in the WeatherBench dataset.}
  \label{fig:visual_weather_v}
  % \vspace{-4mm}
\end{figure}

\noindent{\textbf{Single-variable Wind Component Forecasting}} The quantitative results are presented in Table~\ref{tab:single_wind}. The qualitative visualizations of latitude and longitude wind are shown in Figure~\ref{fig:visual_weather_u} and Figure~\ref{fig:visual_weather_v}. Most recurrent-free models outperform the recurrent-based models.

\begin{table}[ht]
\small
\centering
% \vspace{-2mm}
\renewcommand\arraystretch{1.22}
\caption{The performance on the single-variable wind component forecasting in WeatherBench.}
\resizebox{\textwidth}{!}{%
  \begin{tabular}{ccccccccc}
    \toprule
    \multicolumn{2}{c}{Method} & Params (M) & FLOPs (G) & FPS & MSE $\downarrow$ & MAE $\downarrow$ & RMSE $\downarrow$  \\ \hline
    \rowcolor[HTML]{FDF0E2}
    & ConvLSTM & 15.0 & 136.0 & 43 & 1.8976 & 0.9215 & 1.3775 \\
    \rowcolor[HTML]{FDF0E2}
    & PredRNN & 23.7 & 279.0 & 21 & 1.8810 & 0.9068, & 1.3715 \\
    \rowcolor[HTML]{FDF0E2}
    & PredRNN++ & 38.4 & 414.0 & 14 & 1.8727 & 0.9019 & 1.3685 \\
    \rowcolor[HTML]{FDF0E2}
    & MIM & 37.8 & 109.0 & 122 & 3.1399 & 1.1837 & 1.7720 \\
    \rowcolor[HTML]{FDF0E2}
    & PhyDNet & 3.1 & 36.8 & 172 & 16.7983 & 2.9208 & 4.0986 \\
    \rowcolor[HTML]{FDF0E2}
    & MAU & 5.5 & 39.6 & 233 & 1.9001 & 0.9194 & 1.3784 \\
    \rowcolor[HTML]{FDF0E2}
    \multirow{-7}{*}{\cellcolor[HTML]{FDF0E2}Recurrent-based} & PredRNNv2 & 23.7 & 280.0 & 21 & 2.0072 & 0.9413 & 1.4168 \\
    \rowcolor[HTML]{E7ECE4}
    & SimVP & 14.7 & 8.0 & 430 & 1.9993 & 0.9510 & 1.4140 \\
    \rowcolor[HTML]{E7ECE4}
    & TAU & 12.2 & 6.7 & 505 & 1.5925 & 0.8426 & 1.2619 \\
    \rowcolor[HTML]{E7ECE4}
    & SimVPv2 & 12.8 & 7.0 & 529 & 1.5069 & 0.8142 & 1.2276 \\
    \rowcolor[HTML]{E7ECE4}
    & ViT & 12.4 & 8.0 & 427 & 1.6262 & 0.8438 & 1.2752 \\
    \rowcolor[HTML]{E7ECE4}
    & Swin Transformer & 12.4 & 6.9 & 559 & \textbf{1.4996} & 0.8145 & \textbf{1.2246} \\
    \rowcolor[HTML]{E7ECE4}
    & Uniformer & 12.0 & 7.5 & 466 & 1.4850 & \textbf{0.8085} & 1.2186 \\
    \rowcolor[HTML]{E7ECE4}
    & MLP-Mixer & 11.1 & 5.9 & 687 & 1.6066 & 0.8395 & 1.2675 \\
    \rowcolor[HTML]{E7ECE4}
    & ConvMixer & 1.1 & 1.0 & 1807 & 1.7067 & 0.8714 & 1.3064 \\
    \rowcolor[HTML]{E7ECE4}
    & Poolformer & 10.0 & 5.6 & 746 & 1.6123 & 0.8410 & 1.2698 \\
    \rowcolor[HTML]{E7ECE4}
    & ConvNext & 10.1 & 5.7 & 720 & 1.6914 & 0.8698 & 1.3006 \\
    \rowcolor[HTML]{E7ECE4}
    & VAN & 12.2 & 6.7 & 549 & 1.5958 & 0.8371 & 1.2632 \\
    \rowcolor[HTML]{E7ECE4}
    & HorNet & 12.4 & 6.9 & 539 & 1.5539 & 0.8254 & 1.2466 \\
    \rowcolor[HTML]{E7ECE4}
    \multirow{-13}{*}{\cellcolor[HTML]{E7ECE4}Recurrent-free} & MogaNet & 12.8 & 7.0 & 441 & 1.6072 & 0.8451 & 1.2678 \\
    \bottomrule
  \end{tabular}%
}
% \vspace{-6mm}
\label{tab:single_wind}
\end{table}

% \newpage

\begin{figure}[ht]
  % \vspace{-4mm}
  \centering
  \includegraphics[width=1.0\textwidth]{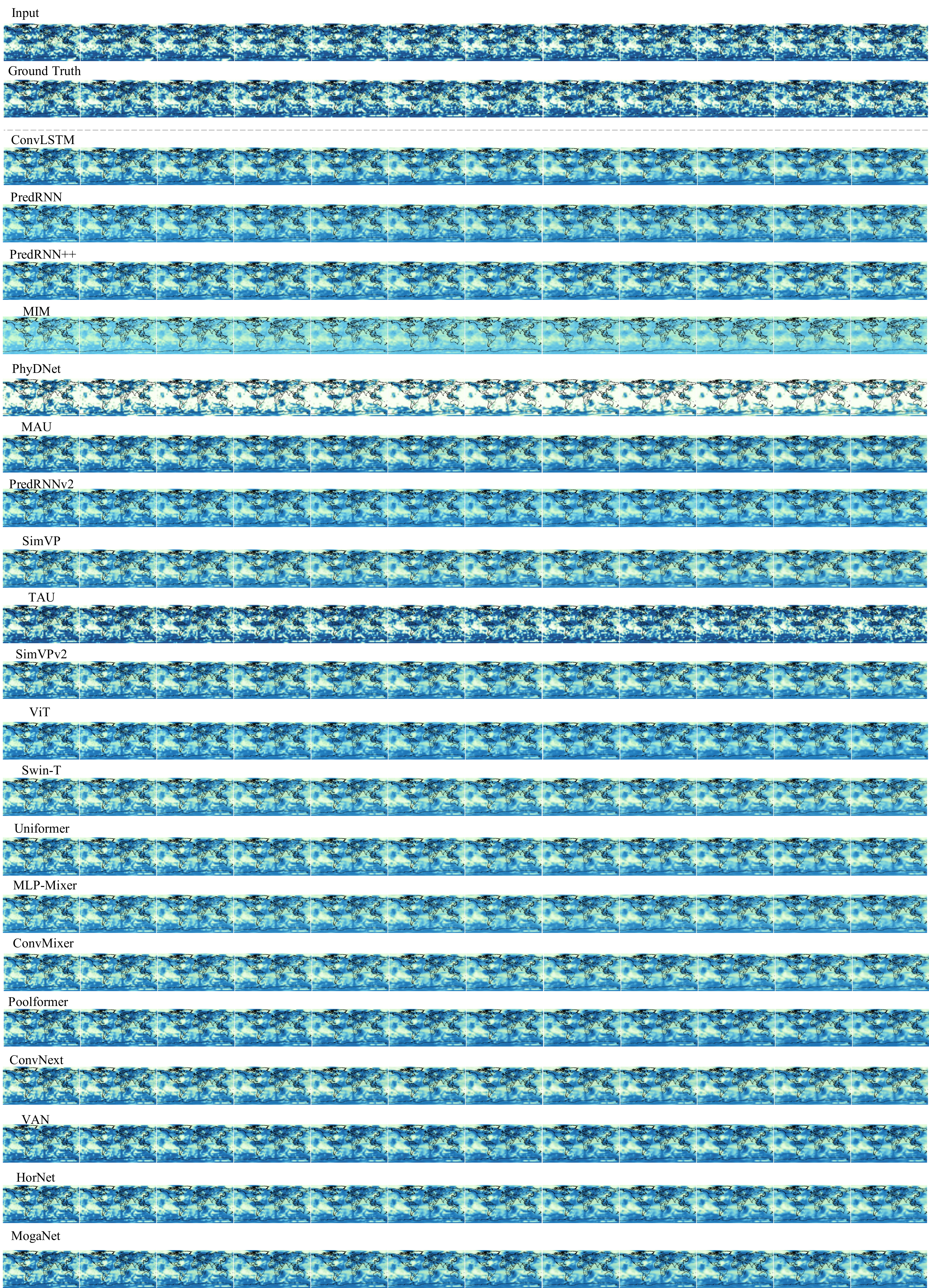}
  \caption{The qualitative visualization on the single-variable cloud cover forecasting in the WeatherBench dataset.}
  \label{fig:visual_weather_tcc}
  % \vspace{-4mm}
\end{figure}

\noindent{\textbf{Single-variable Cloud Cover Forecasting}} The quantitative results and visualization are presented in Table~\ref{tab:single_cloud} and Figure~\ref{fig:visual_weather_tcc}. All the recurrent-free models perform better than their counterparts.

\begin{table}[h]
\small
\centering
% \vspace{-2mm}
\renewcommand\arraystretch{1.22}
\caption{The performance on the single-variable cloud cover forecasting in WeatherBench.}
\resizebox{\textwidth}{!}{%
  \begin{tabular}{ccccccccc}
    \toprule
    \multicolumn{2}{c}{Method} & Params (M) & FLOPs (G) & FPS & MSE $\downarrow$ & MAE $\downarrow$ & RMSE $\downarrow$  \\ \hline
    \rowcolor[HTML]{FDF0E2}
    & ConvLSTM & 14.9 & 136.0 & 46 & 0.04944 & 0.15419 & 0.222 \\
    \rowcolor[HTML]{FDF0E2}
    & PredRNN & 23.6 & 278.0 & 22 & 0.05504 & 0.15877 & 0.234 \\
    \rowcolor[HTML]{FDF0E2}
    & PredRNN++ & 38.3 & 413.0 & 15 & 0.05479 & 0.15435 & 0.234 \\
    \rowcolor[HTML]{FDF0E2}
    & MIM & 37.75 & 109.0 & 126 & 0.05997 & 0.17184 & 0.245 \\
    \rowcolor[HTML]{FDF0E2}
    & PhyDNet & 3.1 & 36.8 & 177 & 0.09913 & 0.22614 & 0.314 \\
    \rowcolor[HTML]{FDF0E2}
    & MAU & 5.5 & 39.6 & 237 & 0.04955 & 0.15158 & 0.222 \\
    \rowcolor[HTML]{FDF0E2}
    \multirow{-7}{*}{\cellcolor[HTML]{FDF0E2}Recurrent-based} & PredRNNv2 & 23.6 & 279.0 & 22 & 0.05051 & 0.15867 & 0.224 \\
    \rowcolor[HTML]{E7ECE4}
    & SimVP & 14.7 & 8.0 & 160 & 0.04765 & 0.15029 & 0.218 \\
    \rowcolor[HTML]{E7ECE4}
    & TAU & 12.2 & 6.7 & 511 & 0.04723 & \textbf{0.14604} & 0.217 \\
    \rowcolor[HTML]{E7ECE4}
    & SimVPv2 & 12.8 & 7.0 & 504 & 0.04657 & 0.14688 & 0.215 \\
    \rowcolor[HTML]{E7ECE4}
    & ViT & 12.4 & 8.0 & 432 & 0.04778 & 0.15026 & 0.218 \\
    \rowcolor[HTML]{E7ECE4}
    & Swin Transformer & 12.4 & 6.9 & 581 & \textbf{0.04639} & 0.14729 & \textbf{0.215} \\
    \rowcolor[HTML]{E7ECE4}
    & Uniformer & 12.0 & 7.5 & 465 & 0.04680 & 0.14777 & 0.216 \\
    \rowcolor[HTML]{E7ECE4}
    & MLP-Mixer & 11.1 & 5.9 & 713 & 0.04925 & 0.15264 & 0.221 \\
    \rowcolor[HTML]{E7ECE4}
    & ConvMixer & 1.1 & 1.0 & 1705 & 0.04717 & 0.14874 & 0.217 \\
    \rowcolor[HTML]{E7ECE4}
    & Poolformer & 10.0 & 5.6 & 722 & 0.04694 & 0.14884 & 0.216 \\
    \rowcolor[HTML]{E7ECE4}
    & ConvNext & 10.1 & 5.7 & 689 & 0.04742 & 0.14867 & 0.217 \\
    \rowcolor[HTML]{E7ECE4}
    & VAN & 12.2 & 6.7 & 523 & 0.04694 & 0.14725 & 0.216 \\
    \rowcolor[HTML]{E7ECE4}
    & HorNet & 12.4 & 6.8 & 517 & 0.04692 & 0.14751 & 0.216 \\
    \rowcolor[HTML]{E7ECE4}
    \multirow{-13}{*}{\cellcolor[HTML]{E7ECE4}Recurrent-free} & MogaNet & 12.8 & 7.0 & 416 & 0.04699 & 0.14802 & 0.216 \\
    \bottomrule
  \end{tabular}%
}
% \vspace{-6mm}
\label{tab:single_cloud}
\end{table}

\clearpage
\clearpage

\clearpage

\clearpage

\noindent{\textbf{Single-variable Temperature Forecasting with High Resolution}} We perform experiments on high-resolution ($128 \times 256$) temperature forecasting. The quantitative results are presented in Table~\ref{tab:single_temp_high}. SimVPv2 achieves remarkable performance, surpassing the recurrent-based models by large margins.

\begin{table}[ht]
  \vspace{-1mm}
  \small
  \centering
  \renewcommand\arraystretch{1.1}
  \caption{The performance on the single-variable high-resolution ($128 \times 256$) temperature forecasting.}
  \resizebox{\textwidth}{!}{%
    \begin{tabular}{ccccccccc}
      \toprule
      \multicolumn{2}{c}{Method} & Params (M) & FLOPs (G) & FPS & MSE $\downarrow$ & MAE $\downarrow$ & RMSE $\downarrow$  \\ \hline
      \rowcolor[HTML]{FDF0E2}
      & ConvLSTM & 15.0 & 550.0 & 35 & 1.0625 & 0.6517 & 1.0310 \\
      \rowcolor[HTML]{FDF0E2}
      & PredRNN & 23.8 & 1123.0 & 3 & 0.8966 & 0.5869 & 0.9469 \\
      \rowcolor[HTML]{FDF0E2}
      & PredRNN++ & 38.6 & 1663.0 & 2 & 0.8538 & 0.5708 & 0.9240 \\
      \rowcolor[HTML]{FDF0E2}
      & MIM & 42.2 & 1739.0 & 11 & 1.2138 & 0.6857 & 1.1017 \\
      \rowcolor[HTML]{FDF0E2}
      & PhyDNet & 3.1 & 148.0 & 41 & 297.34 & 8.9788 & 17.243 \\
      \rowcolor[HTML]{FDF0E2}
      & MAU & 11.8 & 172.0 & 17 & 1.0031 & 0.6316 & 1.0016 \\
      \rowcolor[HTML]{FDF0E2}
      \multirow{-7}{*}{\cellcolor[HTML]{FDF0E2}Recurrent-based} & PredRNNv2 & 23.9 & 1129.0 & 3 & 1.0451 & 0.6190 & 1.0223 \\
      \rowcolor[HTML]{E7ECE4}
      & SimVP & 14.7 & 128.0 & 27 & 0.8492 & 0.5636 & 0.9215 \\
      \rowcolor[HTML]{E7ECE4}
      & TAU & 12.3 & 36.1 & 94 & 0.8316 & 0.5615 & 0.9119 \\
      \rowcolor[HTML]{E7ECE4}
      & SimVPv2 & 12.8 & 112.0 & 33 & \textbf{0.6499} & \textbf{0.4909} & \textbf{0.8062} \\
      \rowcolor[HTML]{E7ECE4}
      & ViT & 12.5 & 36.8 & 50 & 0.8969 & 0.5834 & 0.9470 \\
      \rowcolor[HTML]{E7ECE4}
      & Swin Transformer & 12.4 & 110.0 & 38 & 0.7606 & 0.5193 & 0.8721 \\
      \rowcolor[HTML]{E7ECE4}
      & Uniformer & 12.1 & 48.8 & 57 & 1.0052 & 0.6294 & 1.0026 \\
      \rowcolor[HTML]{E7ECE4}
      & MLP-Mixer & 27.9 & 94.7 & 49 & 1.1865 & 0.6593 & 1.0893 \\
      \rowcolor[HTML]{E7ECE4}
      & ConvMixer & 1.1 & 15.1 & 117 & 0.8557 & 0.5669 & 0.9250 \\
      \rowcolor[HTML]{E7ECE4}
      & Poolformer & 10.0 & 89.7 & 42 & 0.7983 & 0.5316 & 0.8935 \\
      \rowcolor[HTML]{E7ECE4}
      & ConvNext & 10.1 & 90.5 & 47 & 0.8058 & 0.5406 & 0.8976 \\
      \rowcolor[HTML]{E7ECE4}
      & VAN & 12.2 & 107.0 & 34 & 0.7110 & 0.5094 & 0.8432 \\
      \rowcolor[HTML]{E7ECE4}
      & HorNet & 12.4 & 109.0 & 34 & 0.8250 & 0.5467 & 0.9083 \\
      \rowcolor[HTML]{E7ECE4}
      \multirow{-13}{*}{\cellcolor[HTML]{E7ECE4}Recurrent-free} & MogaNet & 12.8 & 112.0 & 27 & 0.7517 & 0.5232 & 0.8670 \\
      \bottomrule
    \end{tabular}%
  }
  \vspace{-1mm}
  \label{tab:single_temp_high}
\end{table}

\begin{figure}[h]
  % \vspace{-2mm}
  \centering
  \includegraphics[width=1.0\textwidth]{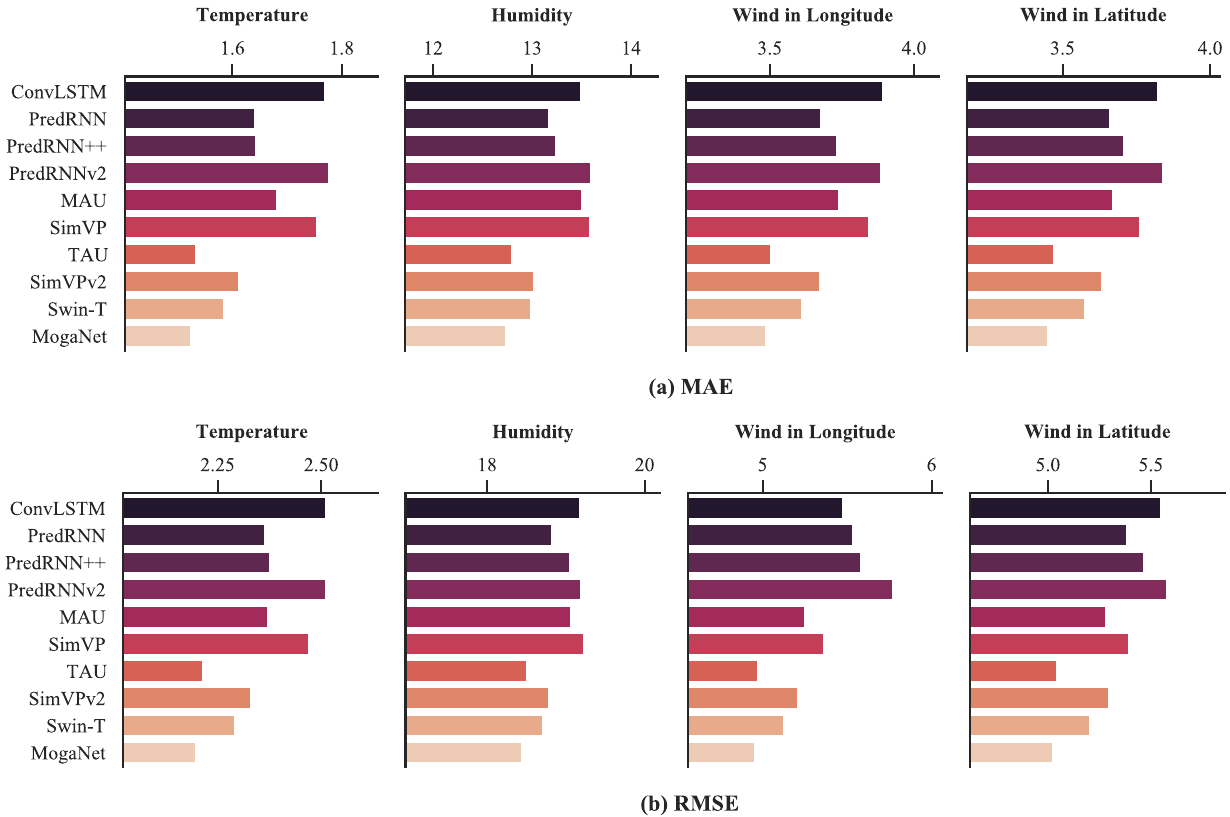}
  \vspace{-3mm}
  \caption{The (a) MAE and (b) RMSE metrics of the representative approaches on the four weather forecasting tasks in WeatherBench (Muti-variable setting).}
  \label{fig:weatherbench_mutli}
  \vspace{-1mm}
\end{figure}

\noindent{\textbf{Multiple-variable Forecasting}} This task focuses on multi-factor climate prediction. We include temperature, humidity, latitude wind, and longitude factors in the forecasting process. The comprehensive results can be found in Table~\ref{tab:multi_temp} to Table~\ref{tab:wind_lon}. We also show a comparison in Figure~\ref{fig:weatherbench_mutli}. MogaNet achieves significant leading performance across various metrics in predicting climatic factors.

\begin{table}[h]
\small
\centering
\renewcommand\arraystretch{1.2}
\caption{The performance on the multiple-variable temperature forecasting in WeatherBench.}
\resizebox{\textwidth}{!}{%
  \begin{tabular}{ccccccccc}
    \toprule
    \multicolumn{2}{c}{Method} & Params (M) & FLOPs (G) & MSE $\downarrow$ & MAE $\downarrow$ & RMSE $\downarrow$  \\ \hline
    \rowcolor[HTML]{FDF0E2}
    & ConvLSTM & 15.5 & 43.3 & 6.3034 & 1.7695 & 2.5107 \\
    \rowcolor[HTML]{FDF0E2}
    & PredRNN & 24.6 & 88.0 & 5.5966 & 1.6411 & 2.3657 \\
    \rowcolor[HTML]{FDF0E2}
    & PredRNN++ & 39.3 & 129.0 & 5.6471 & 1.6433 & 2.3763 \\
    \rowcolor[HTML]{FDF0E2}
    & MIM & 5.5 & 12.1 & 7.5152 & 1.9650 & 2.7414 \\
    \rowcolor[HTML]{FDF0E2}
    & PhyDNet & 3.1 & 11.3 & 95.113 & 6.4749 & 9.7526 \\
    \rowcolor[HTML]{FDF0E2}
    & MAU & 5.5 & 12.1 & 5.6287 & 1.6810 & 2.3725 \\
    \rowcolor[HTML]{FDF0E2}
    \multirow{-7}{*}{\cellcolor[HTML]{FDF0E2}Recurrent-based} & PredRNNv2 & 24.6 & 88.5 & 6.3078 & 1.7770 & 2.5115 \\
    \rowcolor[HTML]{E7ECE4}
    & SimVP & 13.8 & 7.3 & 6.1068 & 1.7554 & 2.4712 \\
    \rowcolor[HTML]{E7ECE4}
    & TAU & 9.6 & 5.0 & 4.9042 & 1.5341 & 2.2145 \\
    \rowcolor[HTML]{E7ECE4}
    & SimVPv2 & 10.0 & 5.3 & 5.4382 & 1.6129 & 2.3319 \\
    \rowcolor[HTML]{E7ECE4}
    & ViT & 9.7 & 6.1 & 5.2722 & 1.6005 & 2.2961 \\
    \rowcolor[HTML]{E7ECE4}
    & Swin Transformer & 9.7 & 5.2 & 5.2486 & 1.5856 & 2.2910 \\
    \rowcolor[HTML]{E7ECE4}
    & Uniformer & 9.5 & 5.9 & 5.1174 & 1.5758 & 2.2622 \\
    \rowcolor[HTML]{E7ECE4}
    & MLP-Mixer & 8.7 & 4.4 & 5.8546 & 1.6948 & 2.4196 \\
    \rowcolor[HTML]{E7ECE4}
    & ConvMixer & 0.9 & 0.5 & 6.5838 & 1.8228 & 2.5659 \\
    \rowcolor[HTML]{E7ECE4}
    & Poolformer & 7.8 & 4.1 & 7.1077 & 1.8791 & 2.6660 \\
    \rowcolor[HTML]{E7ECE4}
    & ConvNext & 7.9 & 4.2 & 6.1749 & 1.7448 & 2.4849 \\
    \rowcolor[HTML]{E7ECE4}
    & VAN & 9.5 & 5.0 & 4.9396 & 1.5390 & 2.2225 \\
    \rowcolor[HTML]{E7ECE4}
    & HorNet & 9.7 & 5.1 & 5.5856 & 1.6198 & 2.3634 \\
    \rowcolor[HTML]{E7ECE4}
    \multirow{-13}{*}{\cellcolor[HTML]{E7ECE4}Recurrent-free} & MogaNet & 10.0 & 5.3 & \textbf{4.8335} & \textbf{1.5246} & \textbf{2.1985} \\
    \bottomrule
  \end{tabular}%
}
\vspace{-8mm}
\label{tab:multi_temp}
\end{table}

% \paragraph{Humidity}

\begin{table}[ht]
\small
\centering
\renewcommand\arraystretch{1.3}
\caption{The performance on the multiple-variable humidity forecasting in WeatherBench.}
\resizebox{\textwidth}{!}{%
  \begin{tabular}{ccccccccc}
    \toprule
    \multicolumn{2}{c}{Method} & Params (M) & FLOPs (G) & MSE $\downarrow$ & MAE $\downarrow$ & RMSE $\downarrow$  \\ \hline
    \rowcolor[HTML]{FDF0E2}
    & ConvLSTM & 15.5 & 43.3 & 368.15 & 13.490 & 19.187 \\
    \rowcolor[HTML]{FDF0E2}
    & PredRNN & 24.6 & 88.0 & 354.57 & 13.169 & 18.830 \\
    \rowcolor[HTML]{FDF0E2}
    & PredRNN++ & 39.3 & 129.0 & 363.15 & 13.246 & 19.056 \\
    \rowcolor[HTML]{FDF0E2}
    & MIM & 41.7 & 35.8 & 408.24 & 14.658 & 20.205 \\
    \rowcolor[HTML]{FDF0E2}
    & PhyDNet & 3.1 & 11.3 & 668.40 & 21.398 & 25.853 \\
    \rowcolor[HTML]{FDF0E2}
    & MAU & 5.5 & 12.1 & 363.36 & 13.503 & 19.062 \\
    \rowcolor[HTML]{FDF0E2}
    \multirow{-7}{*}{\cellcolor[HTML]{FDF0E2}Recurrent-based} & PredRNNv2 & 24.6 & 88.5 & 368.52 & 13.594 & 19.197 \\
    \rowcolor[HTML]{E7ECE4}
    & SimVP & 13.8 & 7.3 & 370.03 & 13.584 & 19.236 \\
    \rowcolor[HTML]{E7ECE4}
    & TAU & 9.6 & 5.0 & 342.63 & 12.801 & \textbf{18.510} \\
    \rowcolor[HTML]{E7ECE4}
    & SimVPv2 & 10.0 & 5.3 & 352.79 & 13.021 & 18.783 \\
    \rowcolor[HTML]{E7ECE4}
    & ViT & 9.7 & 6.1 & 352.36 & 13.056 & 18.771 \\
    \rowcolor[HTML]{E7ECE4}
    & Swin Transformer & 9.7 & 5.2 & 349.92 & \textbf{12.984} & 18.706 \\
    \rowcolor[HTML]{E7ECE4}
    & Uniformer & 9.5 & 5.9 & 351.66 & 12.994 & 18.753 \\
    \rowcolor[HTML]{E7ECE4}
    & MLP-Mixer & 8.7 & 4.4 & 365.48 & 13.408 & 19.118 \\
    \rowcolor[HTML]{E7ECE4}
    & ConvMixer & 0.9 & 0.5 & 381.85 & 13.917 & 19.541 \\
    \rowcolor[HTML]{E7ECE4}
    & Poolformer & 7.8 & 4.1 & 380.18 & 13.908 & 19.498 \\
    \rowcolor[HTML]{E7ECE4}
    & ConvNext & 7.9 & 4.2 & 367.39 & 13.516 & 19.168 \\
    \rowcolor[HTML]{E7ECE4}
    & VAN & 9.5 & 5.0 & 343.61 & 12.790 & 18.537 \\
    \rowcolor[HTML]{E7ECE4}
    & HorNet & 9.7 & 5.1 & 353.02 & 13.024 & 18.789 \\
    \rowcolor[HTML]{E7ECE4}
    \multirow{-13}{*}{\cellcolor[HTML]{E7ECE4}Recurrent-free} & MogaNet & 10.0 & 5.3 & \textbf{340.06} & 12.738 & 18.441 \\
    \bottomrule
  \end{tabular}%
}
\label{tab:multi_humidity}
\end{table}

% \paragraph{Wind in Latitude}

\begin{table}[ht]
  \small
  \centering
  \renewcommand\arraystretch{1.3}
  \caption{The performance on the multiple-variable latitude wind forecasting in WeatherBench.}
  \resizebox{\textwidth}{!}{%
    \begin{tabular}{ccccccccc}
      \toprule
      \multicolumn{2}{c}{Method} & Params (M) & FLOPs (G) & MSE $\downarrow$ & MAE $\downarrow$ & RMSE $\downarrow$  \\ \hline
      \rowcolor[HTML]{FDF0E2}
      & ConvLSTM & 15.5 & 43.3 & 30.789 & 3.8238 & 5.5488 \\
      \rowcolor[HTML]{FDF0E2}
      & PredRNN & 24.6 & 88.0 & 28.973 & 3.6617 & 5.3827 \\
      \rowcolor[HTML]{FDF0E2}
      & PredRNN++ & 39.3 & 129.0 & 29.872 & 3.7067 & 5.4655 \\
      \rowcolor[HTML]{FDF0E2}
      & MIM & 41.7 & 35.8 & 36.464 & 4.2066 & 6.0386 \\
      \rowcolor[HTML]{FDF0E2}
      & PhyDNet & 3.1 & 11.3 & 54.389 & 5.1996 & 7.3749 \\
      \rowcolor[HTML]{FDF0E2}
      & MAU & 5.5 & 12.1 & 27.929 & 3.6700 & 5.2848 \\
      \rowcolor[HTML]{FDF0E2}
      \multirow{-7}{*}{\cellcolor[HTML]{FDF0E2}Recurrent-based} & PredRNNv2 & 24.6 & 88.5 & 31.120 & 3.8406 & 5.5785 \\
      \rowcolor[HTML]{E7ECE4}
      & SimVP & 13.8 & 7.3 & 29.094 & 3.7614 & 5.3939 \\
      \rowcolor[HTML]{E7ECE4}
      & TAU & 9.6 & 5.0 & 25.456 & 3.4723 & 5.0454 \\
      \rowcolor[HTML]{E7ECE4}
      & SimVPv2 & 10.0 & 5.3 & 28.058 & 3.6335 & 5.2970 \\
      \rowcolor[HTML]{E7ECE4}
      & ViT & 9.66 & 6.12 & 27.381 & 3.6068 & 5.2327 \\
      \rowcolor[HTML]{E7ECE4}
      & Swin Transformer & 9.7 & 5.2 & 27.097 & 3.5777  & 5.2055 \\
      \rowcolor[HTML]{E7ECE4}
      & Uniformer & 9.5 & 5.9 & 26.799 & 3.5676 & 5.1768 \\
      \rowcolor[HTML]{E7ECE4}
      & MLP-Mixer & 8.7 & 4.4 & 30.014 & 3.7840 & 5.4785 \\
      \rowcolor[HTML]{E7ECE4}
      & ConvMixer & 0.9 & 0.5 & 31.609 & 3.9104 & 5.6222 \\
      \rowcolor[HTML]{E7ECE4}
      & Poolformer & 7.8 & 4.1 & 35.161 & 4.0764 & 5.9296 \\
      \rowcolor[HTML]{E7ECE4}
      & ConvNext & 7.9 & 4.2 & 31.326 & 3.8435 & 5.5969 \\
      \rowcolor[HTML]{E7ECE4}
      & VAN & 9.5 & 5.0 & 25.720 & 3.4858 & 5.0715 \\
      \rowcolor[HTML]{E7ECE4}
      & HorNet & 9.7 & 5.1 & 30.028 & 3.7148 & 5.4798 \\
      \rowcolor[HTML]{E7ECE4}
      \multirow{-13}{*}{\cellcolor[HTML]{E7ECE4}Recurrent-free} & MogaNet & 10.0 & 5.3 & \textbf{25.232} & \textbf{3.4509} & \textbf{5.0231} \\
      \bottomrule
    \end{tabular}%
  }
  \label{tab:wind_lat}
\end{table}

% \paragraph{Wind in Longitude}

\begin{table}[ht]
\small
\centering
\renewcommand\arraystretch{1.3}
\caption{The performance on the multiple-variable longitude wind forecasting in WeatherBench.}
\resizebox{\textwidth}{!}{%
  \begin{tabular}{ccccccccc}
    \toprule
    \multicolumn{2}{c}{Method} & Params (M) & FLOPs (G) & MSE $\downarrow$ & MAE $\downarrow$ & RMSE $\downarrow$  \\ \hline
    \rowcolor[HTML]{FDF0E2}
    & ConvLSTM & 15.5 & 43.3 & 30.002 & 3.8923 & 5.4774 \\
    \rowcolor[HTML]{FDF0E2}
    & PredRNN & 24.6 & 88.0 & 27.484 & 3.6776 & 5.2425 \\
    \rowcolor[HTML]{FDF0E2}
    & PredRNN++ & 39.3 & 129.0 & 28.396 & 3.7322 & 5.3288 \\
    \rowcolor[HTML]{FDF0E2}
    & MIM & 41.7 & 35.8 & 35.586 & 4.2842 & 5.9654 \\
    \rowcolor[HTML]{FDF0E2}
    & PhyDNet & 3.1 & 11.3 & 97.424 & 7.3637 & 9.8704 \\
    \rowcolor[HTML]{FDF0E2}
    & MAU & 5.5 & 12.1 & 27.582 & 3.7409 & 5.2519 \\
    \rowcolor[HTML]{FDF0E2}
    \multirow{-7}{*}{\cellcolor[HTML]{FDF0E2}Recurrent-based} & PredRNNv2 & 24.6 & 88.5 & 29.833 & 3.8870 & 5.4620 \\
    \rowcolor[HTML]{E7ECE4}
    & SimVP & 13.8 & 7.3 & 28.782 & 3.8435 & 5.3649 \\
    \rowcolor[HTML]{E7ECE4}
    & TAU & 9.6 & 5.0 & 24.719 & 3.5060 & 4.9719 \\
    \rowcolor[HTML]{E7ECE4}
    & SimVPv2 & 10.0 & 5.3 & 27.166 & 3.6747 & 5.2121 \\
    \rowcolor[HTML]{E7ECE4}
    & ViT & 9.7 & 6.1 & 26.595 & 3.6472 & 5.1570 \\
    \rowcolor[HTML]{E7ECE4}
    & Swin Transformer & 9.7 & 5.2 & 26.292 & 3.6133  & 5.1276 \\
    \rowcolor[HTML]{E7ECE4}
    & Uniformer & 9.5 & 5.9 & 25.994 & 3.6069 & 5.0985 \\
    \rowcolor[HTML]{E7ECE4}
    & MLP-Mixer & 8.7 & 4.4 & 29.242 & 3.8407 & 5.4076 \\
    \rowcolor[HTML]{E7ECE4}
    & ConvMixer & 0.9 & 0.5 & 30.983 & 3.9949 & 5.5662 \\
    \rowcolor[HTML]{E7ECE4}
    & Poolformer & 7.8 & 4.1 & 33.757 & 4.1280 & 5.8101 \\
    \rowcolor[HTML]{E7ECE4}
    & ConvNext & 7.9 & 4.2 & 29.764 & 3.8688 & 5.4556 \\
    \rowcolor[HTML]{E7ECE4}
    & VAN & 9.5 & 5.0 & 24.991 & 3.5254 & 4.9991 \\
    \rowcolor[HTML]{E7ECE4}
    & HorNet & 9.7 & 5.1 & 28.192 & 3.7142 & 5.3096 \\
    \rowcolor[HTML]{E7ECE4}
    \multirow{-13}{*}{\cellcolor[HTML]{E7ECE4}Recurrent-free} & MogaNet & 10.0 & 5.3 & \textbf{24.535} & \textbf{3.4882} & \textbf{4.9533} \\
    \bottomrule
  \end{tabular}%
}
\label{tab:wind_lon}
\end{table}

\end{document}